\documentclass[journal]{IEEEtran}
\ifCLASSINFOpdf
\else
\fi

\usepackage{cite}
\usepackage{amsmath,amssymb,amsfonts}
\usepackage{algorithmic}
\usepackage{graphicx}
\usepackage{textcomp}
\usepackage{xcolor}
\usepackage{subfigure}
\usepackage{multirow}
\usepackage{float}
\usepackage{flushend}

\usepackage[linesnumbered,ruled,vlined]{algorithm2e}

\begin{document}

\title{Deep Adaptive Fuzzy Clustering for Evolutionary Unsupervised Representation Learning}

\author{Dayu~Tan,
        Zheng~Huang,
        Xin~Peng, ~\IEEEmembership{Member,~IEEE},
        Weimin~Zhong,
        and Vladimir~Mahalec
\thanks{This work was supported in part by the National Natural Science Foundation of China  under Grant 61925305, 61890930-3, and Basic Science Center Program 61988101, in part by the International (Regional) Cooperation and Exchange Project under Grant 61720106008. (\emph{Corresponding author: Weimin Zhong; Vladimir Mahalec})}
\thanks{D. Tan is with the Key Laboratory of Advanced Control and Optimization for Chemical Processes, Ministry of Education, East China University of Science and Technology, Shanghai 200237, China, and also with the School of Engineering Practice and Technology, McMaster University, Hamilton, ON L8S 4L7, Canada (e-mail: tandayu19@163.com).}

\thanks{Z. Huang is with the School of Engineering Practice and Technology, McMaster University, Hamilton, ON L8S 4L7, Canada (e-mail: huangz55@mcmaster.ca).}

\thanks{X. Peng is with the Key Laboratory of Advanced Control and Optimization for Chemical Processes, Ministry of Education, East China University of Science and Technology, Shanghai 200237, China (e-mail: xinpeng@ecust.edu.cn).}

\thanks{W. Zhong is with the Key Laboratory of Advanced Control and Optimization for Chemical Processes, Ministry of Education, East China University of Science and Technology, 200237 Shanghai, China, and also with Shanghai Institute of Intelligent Science and Technology, Tongji University, Shanghai, 200092, China (e-mail: wmzhong@ecust.edu.cn).}

\thanks{V. Mahalec is with the School of Engineering Practice and Technology, and also with the Department of Chemical Engineering, McMaster University, ON, L8S 4L7 Hamilton, Canada (e-mail: mahalec@mcmaster.ca).}
}


\maketitle
\begin{abstract}
Cluster assignment of large and complex images is a crucial but challenging task in pattern recognition and computer vision. In this study, we explore the possibility of employing fuzzy clustering in a deep neural network framework. Thus, we present a novel evolutionary unsupervised learning representation model with iterative optimization. It implements the deep adaptive fuzzy clustering (DAFC) strategy that learns a convolutional neural network classifier from given only unlabeled data samples. DAFC consists of a deep feature quality-verifying model and a fuzzy clustering model, where deep feature representation learning loss function and embedded fuzzy clustering with the weighted adaptive entropy is implemented. We joint fuzzy clustering to the deep reconstruction model, in which fuzzy membership is utilized to represent a clear structure of deep cluster assignments and jointly optimize for the deep representation learning and clustering. Also, the joint model evaluates current clustering performance by inspecting whether the re-sampled data from estimated bottleneck space have consistent clustering properties to progressively improve the deep clustering model. Comprehensive experiments on a variety of datasets show that the proposed method obtains a substantially better performance for both reconstruction and clustering quality when compared to the other state-of-the-art deep clustering methods, as demonstrated with the in-depth analysis in the extensive experiments.
\end{abstract}


%
\IEEEpeerreviewmaketitle

\section{Introduction}
%
%
%
%
\IEEEPARstart{C}{lustering} is a kind of unsupervised learning method and a fundamental data analysis tool that has been extensively studied and applied to different fields [1]. This type of method iteratively groups the features with a standard division method, for example, k-means [2] and fuzzy means [3], which use the subsequent assignment as supervision to update the clusters. Numerous different clustering methods are calculated by distance functions, and their combinations with embedding algorithms have been explored in literature [4, 5]. The conventional method always employs Euclidean distance to express this similarity, while stochastic neighbor embedding (SNE) converts this distance relationship into a conditional probability to express similarity. In this way, the distance between similar sample points in high-dimensional space is also similar to that in low-dimensional space. They are just focusing on the distance function of the feature space in which to perform clustering. However, clustering high-dimensional and large-scale datasets are complicated due to interpoint distances for each sample become less informative in high-dimensional spaces. It is challenging for these traditional clustering methods to deal with huge and complex datasets [6]. 

Moreover, despite the primeval success of clustering approaches in unsupervised learning, such as image classification [7, 8], very few works [9, 10] have been proposed to employ them to the end-to-end training of ConvNets, and rarely at large scale. In this way, if given a huge collection of unlabeled images represented by raw pixels, how to partition them into a known number of clusters in terms of latent semantics via clustering? A brand of artificial intelligence, known as deep learning, has been proven so useful applied in many fields with its powerful expression ability [11, 12]. Many novel clustering methods that joint in deep embedding frameworks or employ deep neural network models have been presented, we refer to this new type of clustering algorithms as Deep Clustering. Accordingly, the existing deep clustering methods mainly aim to combine the deep feature learning with traditional clustering methods.

Clustering methods combined into a unified deep feature learning framework can directly partition original complex images with even higher clustering performance [13]. Recently advanced ConvNet model combine with feature learning has been successfully adopted in many applications such as target recognition [14] and image classification [15]. These methods jointly train the network model to learn better features and incorporate clustering loss in the objective function of the ConvNet model, where global assumptions were imposed on the latent feature space. Then use the clustering results to direct the network training that jointly refines the latent features and cluster assignments. 

Motivated by the success of deep feature learning framework, deep Beliefs Networks (DBNs) [16] and hierarchies of sparse auto-encoders [17, 18] which attempt to extract features from input data, like deconvolutional network, greedily construct layers from the image upwards in an unsupervised way. In these studies, each layer consists of an encoder and a decoder with a clustering model. Also, convolution neural networks (CNNs) architecture for deep clustering, in which the variational autoencoder (VAE) is the very powerful unsupervised framework for deep generative learning. The widely used CNNs in deep clustering are stacked autoencoders [19] and structured autoencoders (StructAE), which incorporate graph-based clustering to achieve precise deep representations. They are all employed a multi-stage pipeline that pre-trains unified deep network model with unsupervised learning methods and implements traditional clustering methods for large-scale images as post-processing. Nevertheless, they perform deep clustering in cluster assignments with high dependence on the pre-trained similarity matrix, accordingly the clustering performance not good enough on big datasets. 

Furthermore, Yang \emph{et al}. [9] iteratively learn deep feature representations and cluster assignments with a recurrent framework. Successive operations of this framework in a clustering model are expressed as steps in a recurrent process and stacked on top of representations output through a CNNs. Their model presents promising performance on small complex datasets but may be challenging to large numbers of complex images required for multi-convnets to be competitive. Deep embedding clustering (DEC) [20] is a famous deep network structure that implements the feature representation and clustering model training complex datasets simultaneously. This method works through employing highly confidential samples as supervision that the distribution of samples in each cluster is grouped more densely and clearly. However, there is no guarantee of pulling samples near margins towards the correct cluster when training huge and complex datasets with SGD, which also does not guarantee fast convergence.

Although joint clustering and feature learning of ConvNets methods have shown remarkable performance in unsupervised learning, the training schedule alternating between feature clustering and update of the network parameters leads to unstable learning of feature representations [21]. Also, they do not jointly optimize for the deep representation learning and clustering. Deep convnet models with self-evolution clustering [22] and large-scale many-objective decision clustering [23] have gained increasing attention due to their ability to deal with large scale datasets and complex representations. The platform with powerful computational capabilities is typically required for the application of deep ConvNets structures due to high computation complexity. Some clustering approaches have been investigated with the deep ConvNets, but the critical ingredients for deep clustering remain unclear. For instance, how to efficiently group instances into clusters for huge complex data and also provide effective information that defines clustering oriented loss function? How to implement a good trade-off between accuracy and efficiency, as well as improve the performance of clustering with the Deep ConvNets? What types of neural network structures are proper for feature representation learning of clustering?

In this study, we aim to develop new evolutionary representation learning solutions to the deep unsupervised clustering problems. In summary, the main contributions of this study are as follows:

\begin{enumerate}

\item We propose the DAFC to partition group images automatically, and the resulting iterative optimization problem can be efficiently solved by mini-batch RMSprop and back-propagation rather than SGD, a much more clustering-friendly bottleneck space can be learned.

\item We carefully formulated the objective to encompass three crucial aspects essential for high clustering performance: an effective latent representation space, similarity metric, and a weighted adaptive entropy of deep clustering, which can be integrated into the deep learning architecture.

\item This deep evolutionary unsupervised representation learning for both network loss and reconstruction loss with fuzzy clustering offers superior performance compared to cases where these objectives are optimized separately.

\item In weighted adaptive entropy of deep clustering, we calculate fuzzy membership and optimal weight as global variables that can be jointly optimized for the deep representation learning, efficiently group instances into clusters for huge complex data and implement a good trade-off between accuracy and efficiency, as well as further improve the performance of clustering with the deep ConvNets. 
\end{enumerate}

The remainder of this study is organized as follows. We start with a review of the related works that contain some efficient deep clustering frameworks in Section II. We present the proposed deep adaptive fuzzy clustering in Section III. In Section IV, we discuss the results and distributions of state-of-the-art deep clustering methods and compare them with our method. The conclusions of this study and future works are given in Section V.

\section{Related work}

In recent years, many unsupervised feature representation learning algorithms are based on deep generative or other models, which usually use a latent representation space to extract features and then reconstruct input images [24, 25]. Deep clustering is a combination method that joint deep neural network and clustering in the unsupervised feature representation learning. Compared with the traditional clustering algorithm, deep clustering structure is more complex in which the memory consumption and calculation complexity are higher. Moreover, since the popularity of powerful computing platforms, the deep clustering frameworks have proven to provide better clustering performance compared to traditional clustering algorithms [22]. Especially the deep clustering methods training on huge and complex datasets will obtain good enough clustering results and accuracy by adjusting the hyper-parameters.

In deep feature representation learning, shallow neural network inputs a large number of original data and trains them. When the input data is high-definition pictures, and the amount of input information may reach tens of millions, it is very hard for shallow neural network to learn directly from large information resources, thus resulting in improved accuracy by a fraction of one percent by nearly doubling the number of input information. Therefore, researchers [26, 27] compress the input information and extract the most representative information from the input information. These methods reduce the amount of input information and transform it into training data on the neural network. Autoencoder is an unsupervised feature learning method, which is mainly used for data dimensionality reduction and feature extraction. In experiments [13], deep autoencoders can implement better compression efficiency than shallow autoencoders or linear autoencoders. While the autoencoder method is similar to traditional spectral clustering in theory, the former is much more flexible in jointly additional constraints. 

The general training strategy of Stacked AutoEncoder (SAE) [28] is to train a bunch of shallow AutoEncoder (AE) that greedily pre-train the corresponding deep network architecture; that is, SAE is a combination of multiple shallow autoencoders. Each layer is based on the expression features of the previous layer, extracts more abstract and suitable features, and then implements cluster assignments [29]. At the top of the deep feature representation learning is the AE or sparse Encoder, which are both single AE models. In this type of network, parameters $g$ and $w$ of each Encoder are subject to the loss function of its own layer, so the gradient of the $encoder_1$ parameter depends on the gradient of layer $L_1$. After the training of $encoder_1$, parameters are feedback to $L_1$, and then superior parameters and features are transmitted to $encoder_2$ for the next training iteration. A closed-loop back-propagation of the whole iterative update network is formed in the training process. Clustering combined with SAE to learn a lower dimensional representation space, to obtain effective features used for different types of clustering. Although this deep clustering model is flexible enough to partition the complex input data, it always gets stuck easily in non-optimal local minima and results in undesirable cluster assignments [30]. 

\begin{figure}[htbp]
\centerline{\includegraphics[height=3.71cm, width=8.8cm]{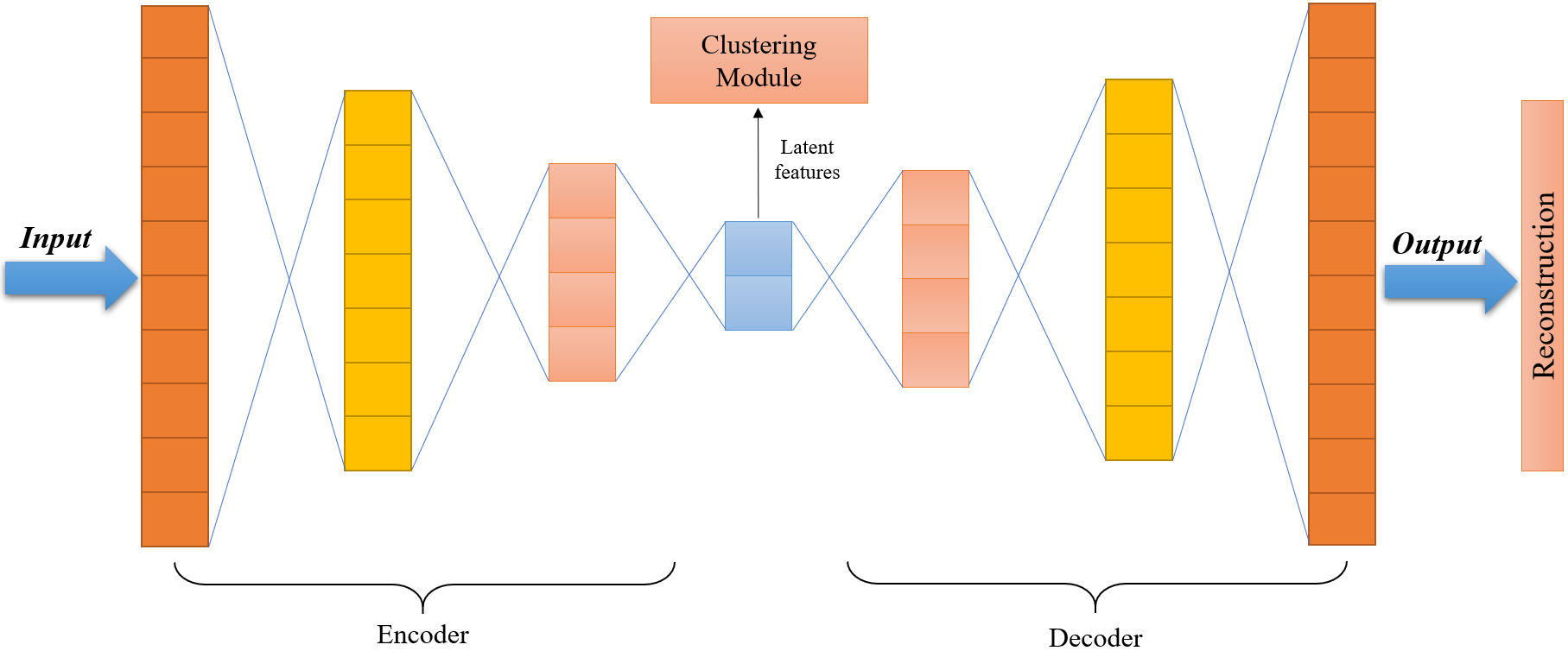}}
\caption{The schematic representation of the variational autoencoder.}
\end{figure}

The network of variational autoencoder (VAE) is composed of encoder and decoder, in which the latent feature layer is followed by the former, as shown in Fig. 1. The encoder maps the input samples to the latent feature layers. The latent feature layer learns the latent features of the input data from encoder model and maps the high-dimensional features to the low-dimensional subspace, and then uses clustering model to partition the mapped data. The decoder of VAE recovers and reconstructs the features so that the feature data can be reconstructed to the original data. It is expected that the input and output of the VAE can be reconstructed in a consistent or lossless manner, and the dimension of latent feature vector is much smaller than that of the input samples. It will be more efficient to use the learned latent layer to cluster assignments and other tasks. This method is conducive to learning more important features and neglecting some redundant features. However, if there is a large difference between the size of the target and the background of the training image, the training network will easily to ignore the target features in the learning process [31]. Therefore, when training huge and complex datasets such as ImageNet dataset, the accuracy of VAE is not high and would not even be able to partition it.

Even though such a scheme takes advantage of SAE and VAE to map the input data into a representative feature space followed by clustering analysis, feature representation space and clustering method are two separate processes, the objective function of which are not optimized jointly. In the next section, we will build deeper feature extraction and reconstruction models based on ConvNets structure, in which we joint the fuzzy clustering model.

\section{Methodology}
In this section, we describe our deep evolutionary unsupervised representation learning, and show that it is possible to obtain useful general-purpose huge and complex features with a deep clustering framework.

\subsection{Network Architecture}

Modern deep network approaches to computer vision, based on statistical learning and deep CNNs, require a good image featurization. In the DAFC method, we build a more in-depth feature extraction and reconstruction model and connect them through latent representation bottleneck layers with convolution operation. Effective latent representation of bottleneck space is a crucial aspect of the deep evolutionary unsupervised representation learning, which can better extract features between each layer of the network. A local structure is constructed in the bottleneck space that joint clustering to achieve better cluster assignments. Comparison with state-of-the-art deep clustering, it is easy to assume that this superiority is due to the fact that bottleneck space can preserve the local structure of input data by minimizing the reconstruction and clustering loss. We design an effective convnet-based classifier with a fuzzy clustering model to extensively exploit the bottleneck space, in which the loss function is the sum of reconstruction loss and fuzzy clustering loss.

The network architecture of DAFC aims to multi-group data points divide into clusters entirely without any labels and jointly optimize for the deep representation learning that further enhances the robustness to the outliers. The mutual information among different types of layers (deep and shallow) for the same sample should be maximized. To improve the accuracy of deep clustering, we increase the number of feature training layers of the network in our model. Inspired by ResNet [32] and DenseNet learning [33], we attempt to control the network optimizer such that Stochastic Gradient Descent (SGD), Adam, and RMSprop are compared. On the gradient descent of the network, we employ a slow learning rate to deal with this bouncing between the ridges problem, which helps obtain a better clustering performance.

The critical factor of the proposed network module is that we utilize convolutional layers with stride and pooling layers instead of convolutional layers followed by the pooling layer in our deep ConvNets model. It is different from the module of the VAE. The modules designed in this work lead to a higher capability of the transformation. We stacked multiple combinations of convolutional layers, the ReLU layer, Batch Normalization, Pooling, and Dropout layer, as shown in Fig. 2. Furthermore, the deep ConvNets model generally reduces the size of the feature maps through Pooling layers with stride $>$ 1. Each module of the designed model connects each layer to every other layer with a feed-forward way, from which module 1 is a feed-forward neural network possessing 22 layers, and module 2, same with former possesses 8 layers. Multiple fully connected layers as the latent representation of bottleneck space are used in this model. We can also flexibly deepen the construction of the network module according to the requirements of input data. When training on enough complex data, this method constantly achieves the best performance on standard competitive classification benchmarks. 

\begin{figure}[htbp]
\centerline{\includegraphics[height=4.1cm, width=9cm]{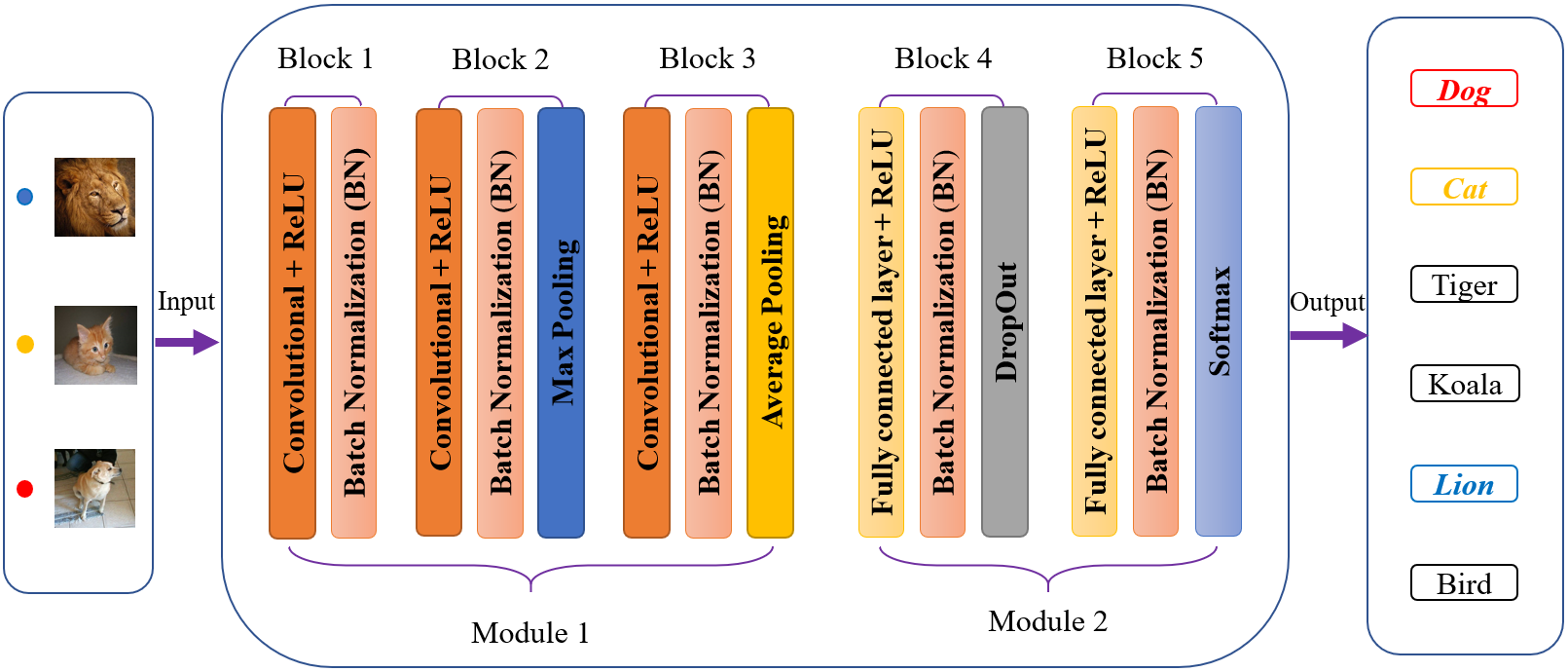}}
\caption{The network modules of the designed model.}
\end{figure}

\begin{figure*}[!t]
\centerline{\includegraphics[height=7.41cm, width=14.65cm]{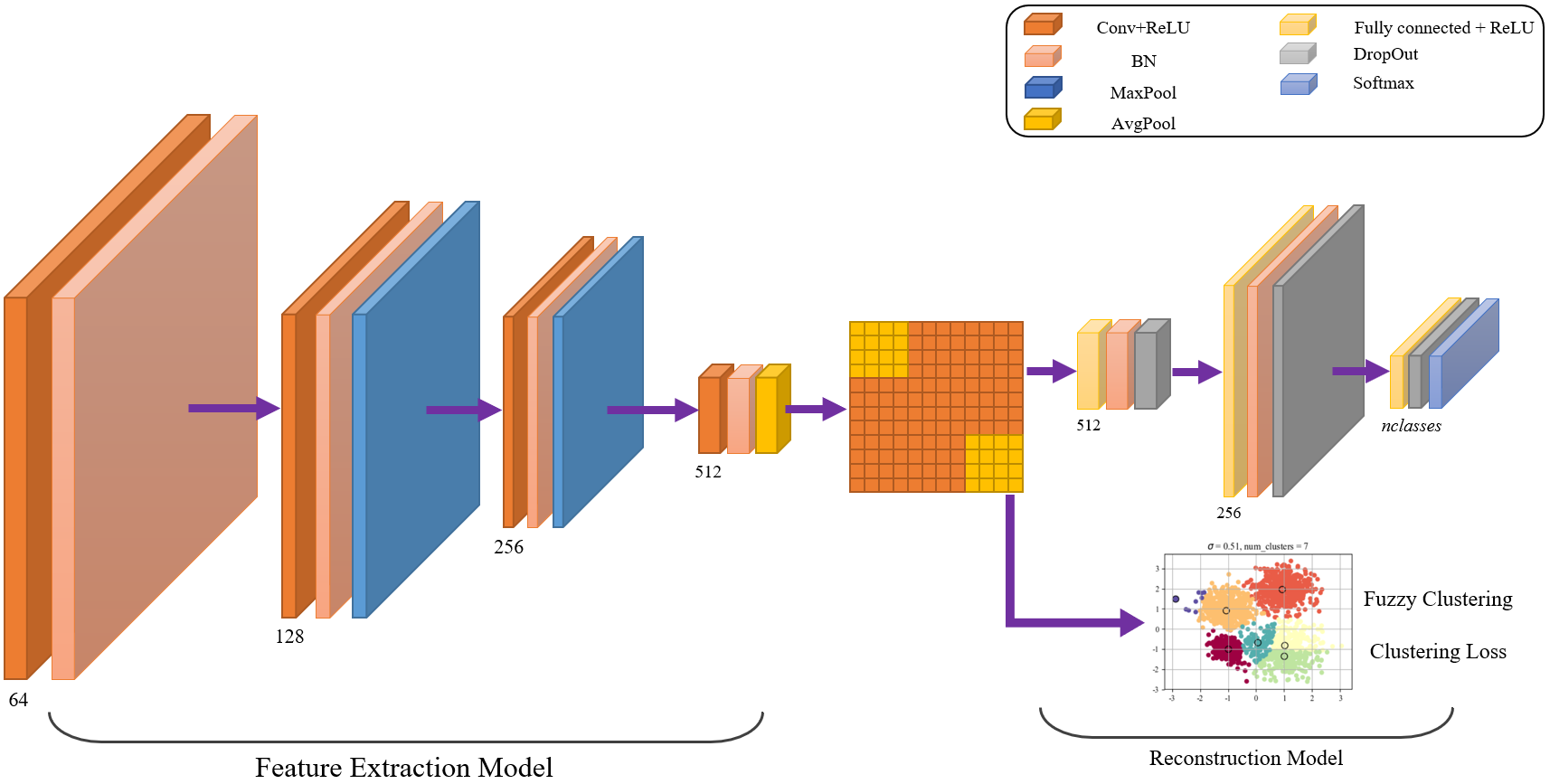}}
\caption{Illustration of overall network architecture.}
\end{figure*}

In our network model, we use multiple fully connected layers that the number of fully connected layers is deepened to improve the model ability of non-linear expression and feature learning. In order to prevent the learning ability of the deepened model from being too good and causing overfitting, we add the dropout layer, followed by the fully connected layers. The dropout layer allows only part of the network parameters to be adjusted for each back-propagation. As shown in Fig. 3, we build the deep ConvNet model for DAFC according to network modules. In these modules, every block has different function layers. Furthermore, Module 1 and Module 2 correspond to each part of the designed network model, respectively.

In this network architecture, Module 1 is designed to be the Feature Extraction model, Module 2 is a Reconstruction model, according to Fig. 2. The feature extraction model (FE) provides a bottom-up mapping from the input to latent representation bottleneck space, while the reconstruction model (Rec) maps the extracted features back to the input space, hopefully giving a reconstruction close to the original data. The feature extraction and reconstruction models and connect them through latent representation bottleneck layers with convolution operation. A local structure is constructed in the bottleneck space that joint fuzzy clustering to achieve better cluster assignments. In the Rec model, we design an effective deep convnet-based classifier with a weighted adaptive entropy fuzzy clustering model to extensively exploit the bottleneck space. Also, we investigate a joint strategy in which the loss function is to minimize the sum of the reconstruction loss and fuzzy clustering loss, where the parameters of FE model $F$ = $f_w$($x$) and Rec model $G$ = $g_\theta$($F$) in the deep ConvNets. 

To this end, we joint the reconstruction loss of feature extraction model and fuzzy clustering loss into objective function simultaneously. The feature extraction model of deep ConvNets will preserve the local structure of complex data generating distribution, avoiding the corruption of feature space. Then, as the learning progresses, the designed deep ConvNets can be employed for measuring more accurate similarities by iterative training, and more cluster assignments will be gradually selected to find more refined groups. The fuzzy clustering model aims to group similar or the same patterns among massive and different data points. In this deep clustering model, we use the membership degree to make the clustering results more discriminative, $\mu$ is the probability of assignment that data point $x$ belongs to the $j$-th cluster. 

\subsection{Deep Clustering Strategy}

In our network architecture, the FE and Rec models, overlap all the features of complex multi-level where the overlapping domains, which are a hierarchical composition of the entire architecture. It can easily be stacked to form a hierarchy by treating the feature maps $f$($x$) of layer $l$ as input for layer $l$ + 1, in which there is an initial non-linear mapping of input data $x$ to the bottleneck space $Z$($x$), as shown in Fig. 4. The mappings $F$ and $G$ are implemented via using the FE and Rec models, and they are connected by the optimized bottleneck space with convolutional and multiple fully connected layers. The bottleneck space representation is improved, such that pairs of mapped points which have a high probability of belonging to the same cluster will be drawn closer together. Therefore, bottleneck space is designed to connect the FE model and the Rec model, which is more meaningful for joint fuzzy clustering.

\begin{figure}[htbp]
\centerline{\includegraphics[height=3.77cm, width=8.8cm]{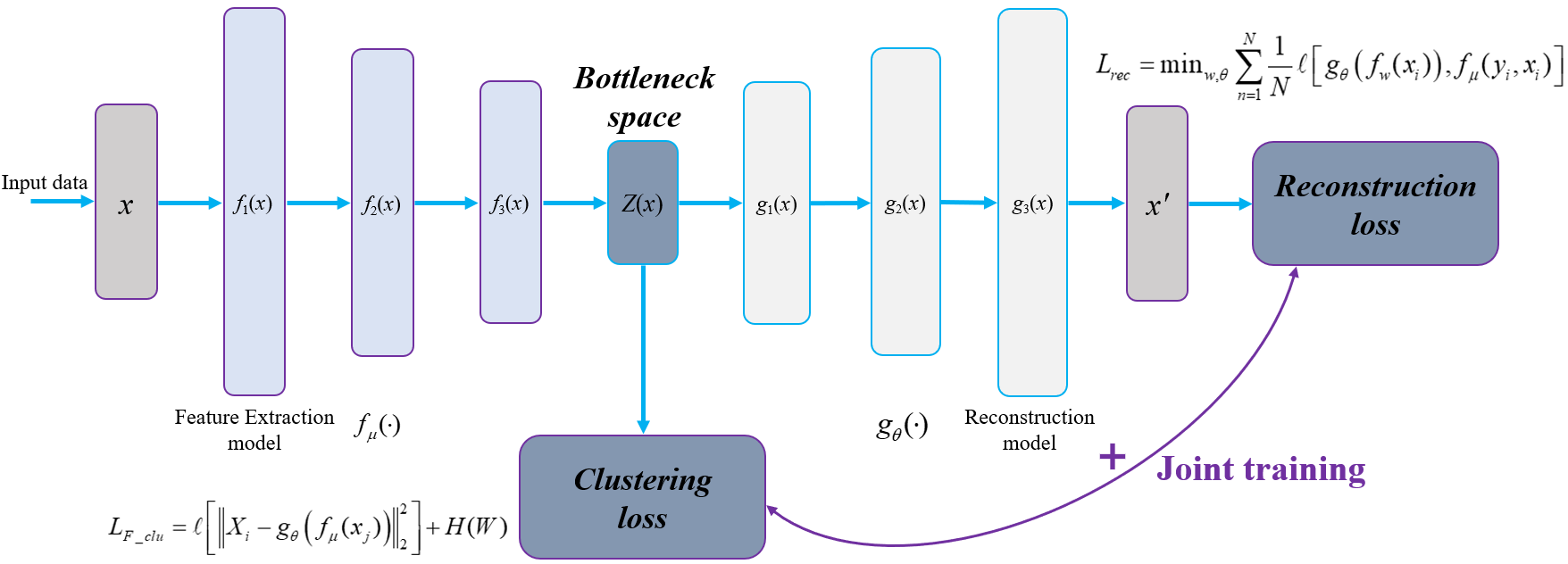}}
\caption{The network modules of the designed model.}
\end{figure}

$f$ is the feature of the input data in the bottleneck space with a fuzzy membership degree $\mu$ during training, which is shown as the similarity structure in Fig. 5, $z$ is a bottleneck space representation. They also represent the result of feature partition by the ConvNets classifier under network hyperparameters. The parameter $\theta$ of the ConvNets classifier and the parameter $w$ of the mapping are jointly learned via optimizing the Loss function of reconstruction that is formulated as follows,

$$L_{R e c} = \min _{w, \theta} \sum_{i = 1}^{M} \frac{1}{M} \ell\left[g_{\theta}\left(f_{w}\left(x_{i}\right)\right), z_{i}\right]. \eqno{(1)}$$

First, we give a batch of original data $\left \{ x \right \}$ plus the transformed data $\left \{ {x}' \right \} $ through data augmentation, then get the input data: $x_{i} = x+{x}'$. $x_i$ represents the $i$-th image and $M$ is the number of images. In the input data, each image $x_i$ is associated with a label $y_i$ in $\left \{ 0,1 \right \} ^{k} $. The network maps the input data into compact feature maps $F$ = $f_w$($x_i$) and clustering the output of the FE model, then use the subsequent cluster assignments as pseudo-labels to optimize the Eq. (1) as well as back-propagate to Rec model as feedback information. The next step generates feature labels of the input data through using the network of Rec model in which the similarity matrix reads the pseudo-labels for this batch from samples memory of a ConvNets. Fig. 5 is a schematic representation of the proposed deep evolutionary unsupervised representation learning with a fuzzy clustering model. We define the similarity matrix in \emph{Definition 1}, and for latent bottleneck space representation $z_i$ = $f_\mu$($y_i$, $x_i$). With the similarity matrix, we update the deep network with the best optimizer to joint the fuzzy clustering model as follows, 

$$L_{R e c}=\min _{w, \theta} \sum_{i=1}^{M} \frac{1}{M} \ell\left[g_{\theta}\left(f_{w}\left(x_{i}\right)\right), f_{\mu}\left(y_{i}, x_{i}\right)\right]. \eqno{(2)}$$
where $f_\mu$($\cdot$) sets a fuzzy membership degree for each column of the similarity matrix, this representation should be more accurate. Based on Eq. (2), we further obtain

$$f_{\mu}\left(y_{i}, x_{i}\right)=C_{i, j}\left(x_{i}\right) \cdot y_{i}. \eqno{(3)}$$

\textbf{\emph{Definition 1}}: Similarity matrix. We assume that $C$ is the adjacency matrix of generate feature labels and $a_i$ is the $i$-th column of $C$. The similarity between node $i$ and node $j$ of the labels is

$$C_{i, j}(\cdot)=\frac{a_{i}^{T} a_{j}}{\left\|a_{i}\right\|\left\|a_{j}\right\|}=\frac{a_{i}^{T} a_{j}}{\sqrt{d_{i}} \sqrt{d_{j}}}. \eqno{(4)}$$

\begin{figure*}[htbp]
\centerline{\includegraphics[height=8cm, width=16.6cm]{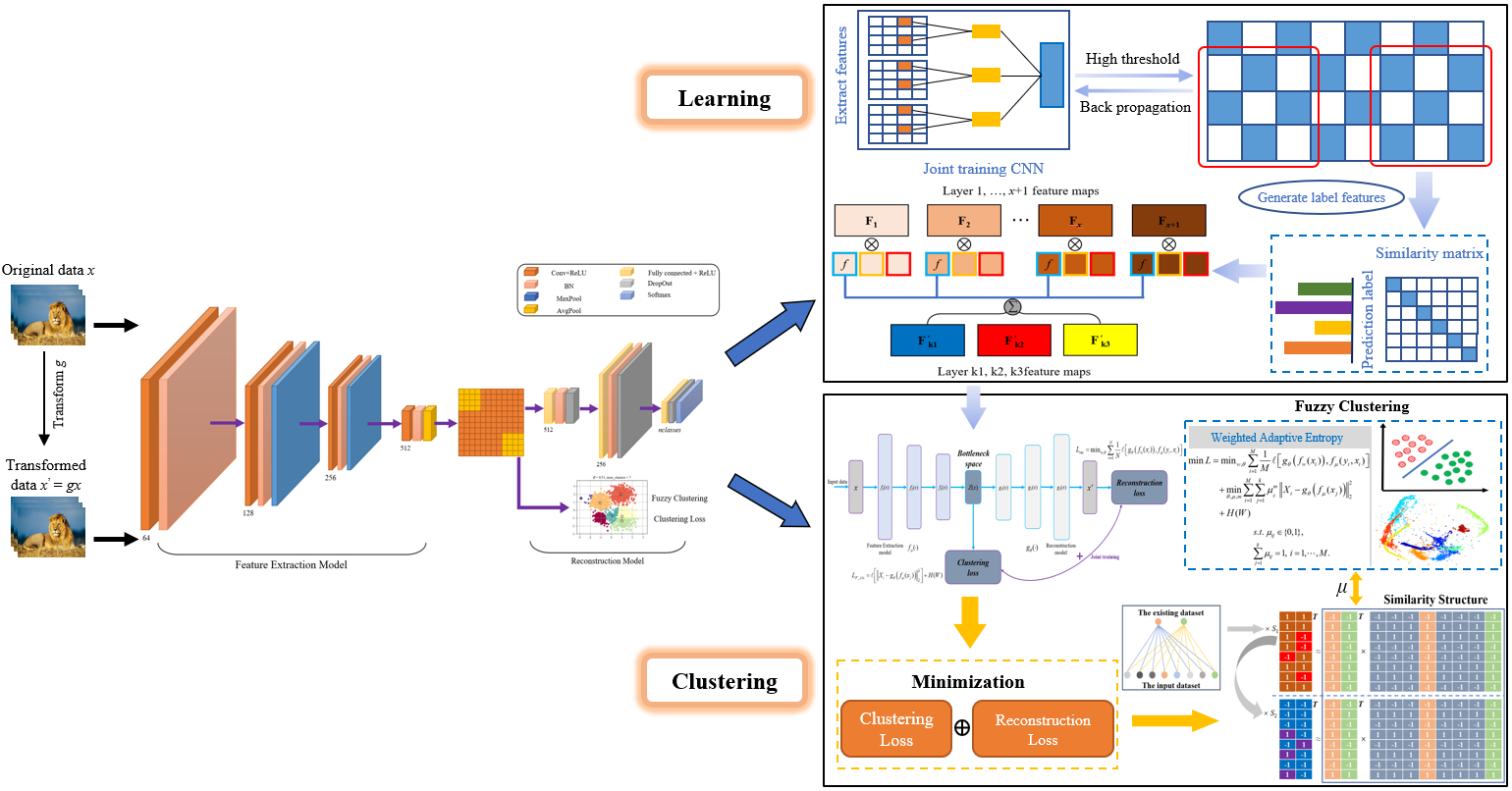}}
\caption{The schematic representation of the proposed DAFC method.}
\end{figure*}

We transfer the features of FE model training to the bottleneck space, in which the fuzzy clustering is added to the objective and optimized along with clustering loss. The results output of the fuzzy clustering model is back-propagated to ConvNets as pseudo labels and optimize the $L_{Rec}$, as well as the network parameters are optimized. After the greedy layer by layer training, we connect all the layers of the FE and Rec model by back transmission to construct a deep joint training model. Then we fine-tune the joint model and set a high threshold and two trade-off parameters to minimize the loss of reconstruction information. The Rec model will preserve the local structure of pseudo labels generating distribution, avoiding the corruption of bottleneck space. This deep clustering model iteratively learns the features of input data and partitions them. 

Also, we calculate the similarity among features and select highly-confident generated label features to feedback the FE and Rec models training through the joint framework. In the Rec model, we set a high-threshold to determine whether some image pseudo labels should be merged into features maps. If the similarity between the two labels is larger than the high-threshold, and we group this type of label belong to the same cluster. Back-propagation for feedback the FE and Rec that aims to learn a deep ConvNets based mapping function $g$, which is parameterized by $\theta$. The parameters of the joint framework are updated via minimizing the loss function of Rec with the fuzzy clustering model,

$$\begin{aligned}
L_{F_{-} c l u} &=\ell\left[\left\|X_{i}-g_{\theta}\left(f_{\mu}\left(x_{j}\right)\right)\right\|_{2}^{2}\right]+H(W) \\
&=\min _{\theta, \mu, m} \sum_{i=1}^{M} \sum_{j=1}^{k} \mu_{i j}^{m}\left\|X_{i}-g_{\theta}\left(f_{\mu}\left(x_{j}\right)\right)\right\|_{2}^{2}+H(W), 
\end{aligned} \eqno{(5)}$$
where $M$ is the number of data samples in the dataset, $j$ is the $j$-th cluster, $k$ is the number of cluster assignments. 

In the loss function of fuzzy clustering, we consider that the weight of a huge and complex dataset is divided into clusters in the bottleneck space, it also represents the probability of contribution of that data feature maps in forming the cluster. We further improve the fuzzy clustering with weighted adaptive loss function by adding weighted entropy that stimulates more features to contribute to clusters identification. In this way, the DAFC model can be trained directly in a deep end-to-end manner even with the weighted adaptive entropy, and this learned hierarchical representation is proved effective for deep clustering tasks.

\textbf{\emph{Definition 2}}: Weighted adaptive entropy of deep clustering. We assume the weight information of the clustering is $H$($w$), in which we set the optimal fuzzy membership degree and optimal weight $w_{ij}$ simultaneously. The $H$($w$) is calculated as follows,

$$H(w)=\lambda_{1}\left(1-\sum_{j=1}^{k} \mu_{i j}\right)+\lambda_{2} \sum_{i=1}^{M} \sum_{j=1}^{k} w_{i j} \log w_{i j}, \eqno{(6)}$$
where $\lambda_1$ is a tradeoff parameter that controls fuzzy robustness of assignments to various type outliers, and $\lambda_2$ also a tradeoff parameter that controls the cluster distribution of fuzzy membership. The objective function of DAFC is

$$\min L=L_{R e c}+L_{F_{-} c l u}, \eqno{(7)}$$ which can be rewritten as:

$$\begin{aligned}
\min L=& \min _{w, \theta} \sum_{i=1}^{M} \frac{1}{M} \ell\left[g_{\theta}\left(f_{w}\left(x_{i}\right)\right), f_{\mu}\left(y_{i}, x_{i}\right)\right] \\
&+\min _{\theta, \mu, m} \sum_{i=1}^{M} \sum_{j=1}^{k} \mu_{i j}^{m}\left\|X_{i}-g_{\theta}\left(f_{\mu}\left(x_{j}\right)\right)\right\|_{2}^{2} \\
&+H(W),
\end{aligned} \eqno{(8)}
$$

$$\begin{aligned}
&\text {s.t. } \mu_{i j} \in\{0,1\},\\
&\sum_{j=1}^{k} \mu_{i j}=1, i=1, \cdots, M.
\end{aligned}$$

Consider the problem of clustering a set of $M$ images $\left \{ x_{1} ,\dots ,x_{M}  \right \}^{k} $ into $k$ clusters in the bottleneck space and each image $x_i$ is associated with a label $y_i$ in $\left\{0, 1 \right \}^{k}$. We also use fuzzy membership $\mu$ represents the image’s probability to one of $k$ possible predefined clusters. The closer the bottleneck space $Z$($x$) of input data $x$ is to the centroid $c_j$, $c_j$ = $g_\theta$($f_\mu$($x_j$)), the higher the fuzzy membership of $x$ belonging to cluster $j$. $\mu$ is the fuzzy membership and $m$ is the Lagrangian multiplier. Then $\mu$ is derived from this objective function as follows,

$$\min \mu=\frac{\partial L}{\partial \mu} \Rightarrow \mu=\left(\frac{\lambda_{1}}{m \cdot \ell}\right)^{\frac{1}{m-1}}, \eqno{(9)}$$
where $\ell=\ell\left[\left\|X_{i}-g_{\theta}\left(f_{\mu}\left(x_{j}\right)\right)\right\|_{2}^{2}\right]$.

We now consider the derivative of function $L$ with the weight at the new step. Given metrics $\mu$ and $c_j$ are fixed, $L$ is minimized for optimal weight $w$ in Eq. (8),

$$w_{i j}=\exp \left[2 \sum_{j=1}^{k} g_{\theta}\left(f_{\mu}\left(x_{j}\right)\right)-1\right]. \eqno{(10)}$$

The addition of adaptive entropy allows clustering to solve the partition problem with a much faster iterative algorithm.

In summary, we have introduced a deep adaptive fuzzy clustering for evolutionary unsupervised representation learning. The pseudo-code of the DAFC is described in Algorithm 1. During each iteration, the deep fuzzy clustering alternately selects similar and dissimilar sample groups via the fixed deep FE and Rec models and trains them based on the selected sample groups. See algorithm 1 for detailed steps,

\begin{algorithm}
  \caption{Deep Adaptive Fuzzy Clustering (DAFC)}
  \KwIn{Dataset $X=\{x_i\}^k_{i=1}$, balance coefficient $\lambda_1$, $\lambda_2$, and high threshold $\epsilon_r$}
  \KwOut{Deep Clustering Results $R$ and $Test\_Err$}
  Randomly initialize $\theta$;
  
  Initialize $C, G,$ and $F$ with pre\_train FE and Rec models;
  
  $\{Acc, ARI, NMI\}$ = $R$;
  
  repeat;
  
  \For{epoch $\in$ $0, 1,$ \dots, MaxEpochs}
  {
  Generate deep ConvNets representations $G$;
  
  \For{iter $\in$ $0, 1,$ \dots, MaxIter}
  {
    float $Max\mu_{ij}=1$\;
    $f_{\mu}(x) \ne null$\;
    
    Calculate the similarity matrix $C$; //Eq. (4)
    
    Calculate $L_{Rec}, L_{F\_clu}$; // Eqs. (2) and (5)
    
    Plus $L_{Rec}$ and $L_{F\_clu}$ by minimizing Eq. (7);
    
    Update $\mu_{ij}$ by minimizing Eq. (9);

    \While{not end of $R$}
    {
      Compute accuracy of the designed ConvNets (1 - $test\_error$)\;
    }
    Update $w_{ij}$ according to Eq. (10);
    
    Back propagation to FE and Rec in DAFC;
    
    Until $\{Acc, ARI, NMI\}$ = Max$R$\;
  $\textbf{end};$
  }
  $\textbf{end};$
  }
  
  return $R, Test\_Err$ \;
\end{algorithm}

\section{ Experiments}
\subsection{ Experiment Setup}

Then, we present a comprehensive comparison with the latest deep clustering models and the traditional clustering methods in Section IV-D. The experiments are all implemented in cudnn v7 and RTX 2080 Ti GPU (NVIDIA) in the Pytorch framework to measure performance and efficiency for different deep clustering methods. We consider the same batch size with the same number of epochs as well as measure inference time in milliseconds. When training on SVHN and STL-10 datasets, we need to pay attention to the amount of memory required for computation, and also carefully balance the Kernel Size and Filter Size if the training is to complete within the desired time frame. For all the convolutional layers, we set different the number of channels (64, 128, 256, and 512) and filter size (2 × 2 and 3 × 3), stride = 1, and padding = 1. For pooling layers, we set kernel size = 2 and stride = 2 in Maxpooling, and we set kernel size = 4 and stride = 4 in Avgpooling. Each layer is pre-trained for 2,000 iterations with a dropout rate of 50\%. We investigate these algorithms on several benchmarks. To the best of our knowledge, we set the max $batch\_size$ for each epoch.

Three optimizers are employed as the optimization algorithm in this work. We have verified the performance when Adam and RMSprop optimizers are employed. Compared to Adam and SGD, we found that Adam suffers from the usual problem of flat-lining and thereby gives a poor solution. SGD was found to require many thousands of iterations for convergence and also give a poor solution. RMSprop leads to more accurate solutions for training deep clustering frameworks on selected complex datasets. In addition, we set the learning rate ($lr$) = 1 × $10^{-4}$. For fair comparison and clear illustration of the effectiveness of the DAFC, the number of training epochs and learning rate for each compared method is kept at the same number. We implement our method and the comparison methods 10 times and present the average results to prevent random cases.

\subsection{Evaluation metrics} 
We use three famous metrics to measure the performance of our model and the comparison model, $Acc$, $ARI$, and $NMI$. One of the metrics, $ARI$, is an improvement of $RI$ that is recommended as the index of choice for measuring agreement between two partitions in clustering analysis with different numbers of clusters. For these metrics, with expected value is between 0 and 1, the closer their values are to 1, and the better the clustering performance. They are computed by as follows:

\noindent$Acc$: Accuracy index of clustering,

$$A c c=\sum_{i=1}^{M} \frac{N_{i}^{j}}{N},$$
where $N^i_i$ is the number of common data samples in the $i$-th object and its matched cluster $j$.

\noindent$ARI$: Adjusted Rand Index,

$$ARI=\frac{ {\textstyle \sum_{ij}\binom{n_{ij} }{2}-\left [ \sum_{i}\binom{a_{i} }{2} \sum_{j}\binom{b_{j} }{2} \right ] /\binom{n}{2}  } }{\frac{1}{2}\left [ \sum_{i}\binom{a_{i} }{2} +\sum_{j}\binom{b_{j} }{2} \right ]-\left [ \sum_{i}\binom{a_{i} }{2} \sum_{j}\binom{b_{j} }{2} \right ]/\binom{n}{2}  }, $$
where $n_{ij}$ represents each element of the $X$ × $Y$ matrix.

\noindent$NMI$: normalized mutual information reflects the quality of clusters, is defined as

$$NMI=2 \frac{I(X ; Y)}{H(X)+H(Y)},$$
where 

$$I(X ; Y)=\sum_{x, y} p(x, y) \log \frac{p(x, y)}{p(x) p(y)},$$
$$H(X)=-\sum_{i=1}^{M} p\left(x_{i}\right) \log _{b} p\left(x_{i}\right).$$

\subsection{Datasets}

In this section, we only consider for comparison with the state-of-the-art and highly competitive datasets, then present thorough experiments demonstrating the efficacy and value of the proposed technique evaluated on a wide range of use cases and deeper structures. We evaluate our network model on six highly competitive complex datasets, which are MNIST, Fashion-MNIST, SVHN, CIFAR-10, CIFAR-100, and STL-10 datasets.

MNIST: It is a very popular and widely used dataset with 70,000 handwritten digits, the input image size of this dataset is set to 28 × 28 pixel size. The 60,000 pattern training set contained examples from approximately 250 writers.

Fashion-MNIST (F-MNIST): This dataset is obtained from different clothes, shoes, and bags images etc., and is a product image dataset that replaces the MNIST handwritten digits. It covers front images of 70,000 different products from 10 categories. The size, format, and training/testing set partition of F-MNIST is completely consistent with the original MNIST. There are 60,000 images in the training set, 10,000 images in the testing set, 10 kinds of the class, and each image belongs to its class. We can directly use it to test the performance of built pattern recognition and deep clustering algorithms.

SVHN: The Street View House Numbers (SVHN) dataset is obtained from house numbers in Google Street View images, which is a real-world image dataset for developing deep learning, data preprocessing, and target recognition algorithms. There are 73,257 digit images for training, 26,032 digit images for testing, and 531,131 additional digits to use as the extra training set.

CIFAR-10: This dataset consists of 60,000 colored images of 10 classes with 32 × 32 pixels; each class has 6,000 images. There are 50,000 training images and 10,000 testing images. This dataset is divided into five training batches and one testing batch, each batch has 10,000 images. The testing batch contains 1,000, randomly selected images from each category. Before training, we first normalize this dataset using the channel means and standard deviations for preprocessing.

CIFAR-100: This dataset is like CIFAR-10, except that it has 100 classes, each class contains 600 images, and each class has 500 images in the training set and 100 images in the testing set. There are 100 classes in CIFAR-100 that are divided into 20 super-classes. Each image has an ``accurate" label (the class it belongs to) and a ``coarse" label (the super-classes it belongs to). During training CIFAR-100, we also first normalize the dataset using the channel means and standard deviations for preprocessing, in which standard deviations are different from CIFAR-10. 

STL-10: It is a dataset of 96 × 96 color images, which is an image recognition dataset for developing unsupervised feature learning and deep learning. It is inspired by the CIFAR dataset and with many modifications. There are 500 images and 10 pre-defined folds in the training dataset, 800 images in the testing dataset, and 100,000 unlabeled images for unsupervised learning. We use all the training data with its data augmentation that increase the length of the training dataset to 4,000.

\subsection{Baselines}

For comparing deep clustering algorithms on these highly competitive datasets, we focus on traditional clustering algorithms, K-means [2], AC [34], and SC [35]; and also advanced network clustering algorithms DeCNN [36], DAE [37], VAE [38], GAN [39], JULE [9], DEC [20], DAC [40], and DCCM [41], which are widely applied in various research and production. We comprehensively demonstrate the effectiveness of the proposed model and compare it with state-of-the-art models on the above datasets, in which we set the same number of epochs as well, and the comparison results are shown in Tables I, II, and III. In the following, we describe our findings with respect to the different deep clustering architectures and analyze the performance of our proposed DAFC.

In this section, we choose each method with the best performance. We compare DAC and DAFC on the MNIST, F-MNIST, and SVHN datasets, due to DAC has the best performance as well as being stable on these three datasets. Since DCCM has the best and stable performance on the other three datasets, we compare DCCM and DAFC on the CIFAR-10, CIFAR-100, and STL-10 datasets. 

\subsubsection{Comparison with DAC}
First, we show for each training and cluster assignment results of the DAC\footnote{https://github.com/vector-1127/DAC} method and our method the following: Acc, ARI, and NMI. By observing and comparing these results, we can obtain which method is better.

\noindent{\textbf{MNIST}}

\begin{figure}[htbp]
\centering
\subfigure[Acc values (DAFC and DAC)]
{
\begin{minipage}[b]{0.2\textwidth}
\centerline{\includegraphics[height=3.4cm, width=9cm]{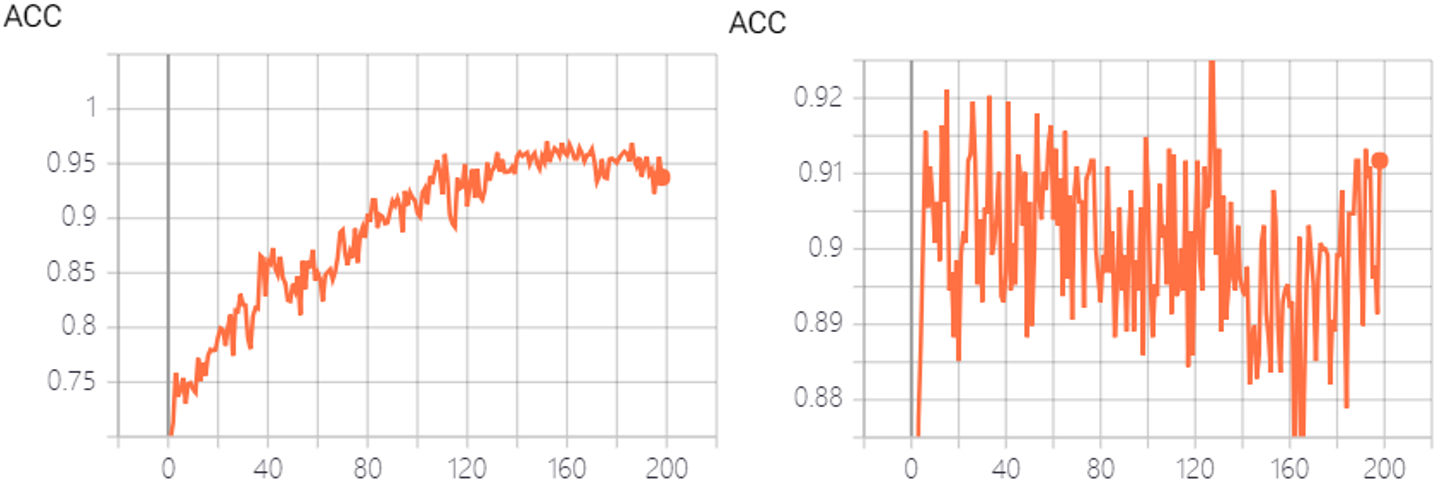}}
\end{minipage}
}

\subfigure[ARI values (DAFC and DAC)]{
\begin{minipage}[b]{0.2\textwidth}
\centerline{\includegraphics[height=3.4cm, width=9cm]{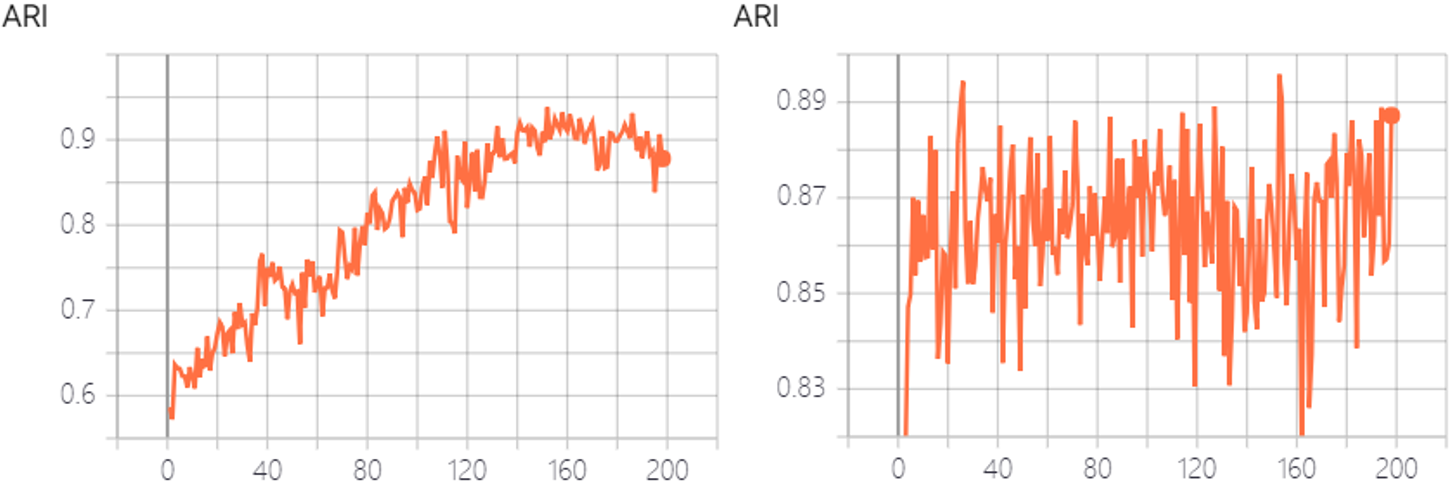}}
\end{minipage}
}

\subfigure[NMI values (DAFC and DAC)]{
\begin{minipage}[b]{0.2\textwidth}
\centerline{\includegraphics[height=3.4cm, width=9cm]{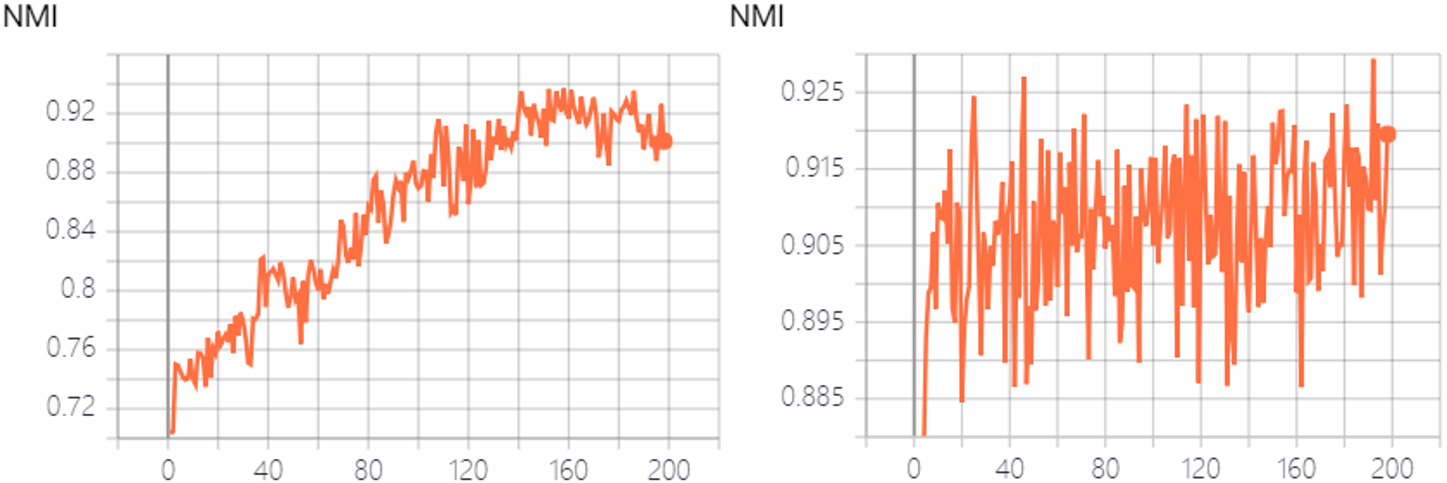}}
\end{minipage}
}

\caption{On the left of the plots (a) to (c) are the performance results of DAFC, which are Acc, ARI, and NMI on the \textbf{MNIST}, respectively. On the right of the plots (a) to (c) are the performance results of DAC.}
\end{figure}

We train DAFC and DAC with optimal parameters on the MNIST dataset. For DAFC, obtaining the top-1 Acc, ARI, and NMI are 0.9703, 0.9380, and 0.9370, respectively. For DAC, obtaining the top-1 Acc, ARI, and NMI are 0.9461, 0.9133, and 0.8944, respectively. Our method improves more than 2.42\%, 2.47\%, and 4.26\% compared to the DAC on MNIST for three metrics. We show each result of cluster assignments for training MNIST in Figs. 6(a), 6(b), and 6(c), in which the x-axis is epochs, and the y-axis is each performance value. From these plots, it can be seen that cluster assignment results of our method gradually increase and achieve the optimal value, and the curve rises smoothly. Compared with the result of DAC oscillation, our method is more stable.

\noindent{\textbf{F-MNIST}}

\begin{figure}[htbp]
\centering
\subfigure[Acc values (DAFC and DAC)]
{
\begin{minipage}[b]{0.2\textwidth}
\centerline{\includegraphics[height=3.4cm, width=9cm]{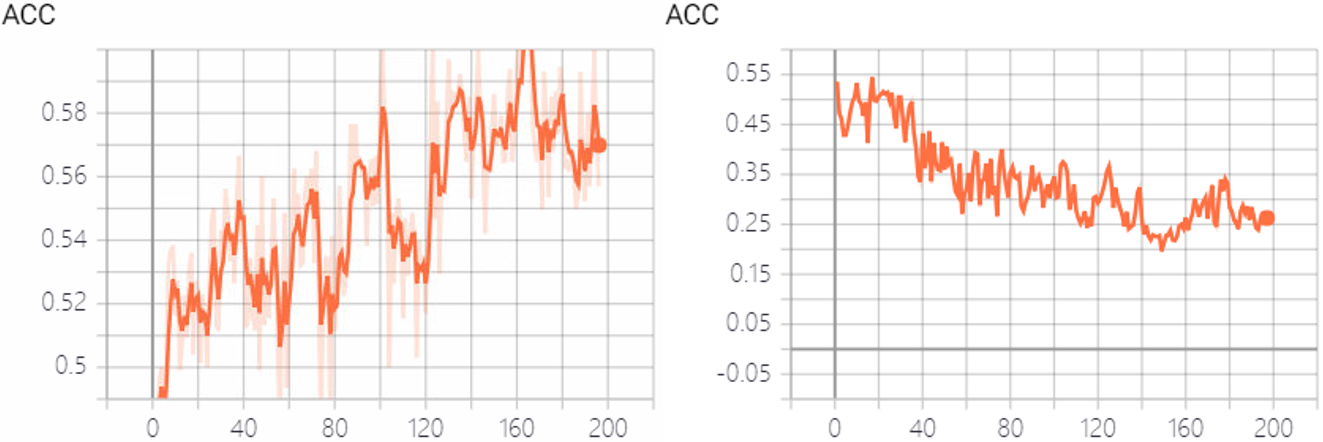}}
\end{minipage}
}

\subfigure[ARI values (DAFC and DAC)]{
\begin{minipage}[b]{0.2\textwidth}
\centerline{\includegraphics[height=3.4cm, width=9cm]{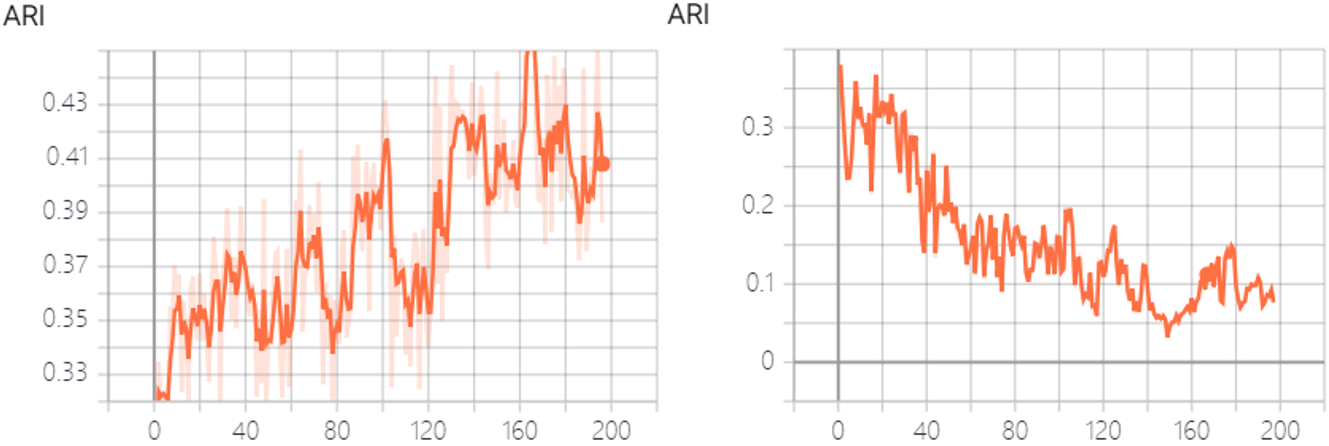}}
\end{minipage}
}

\subfigure[NMI values (DAFC and DAC)]{
\begin{minipage}[b]{0.2\textwidth}
\centerline{\includegraphics[height=3.4cm, width=9cm]{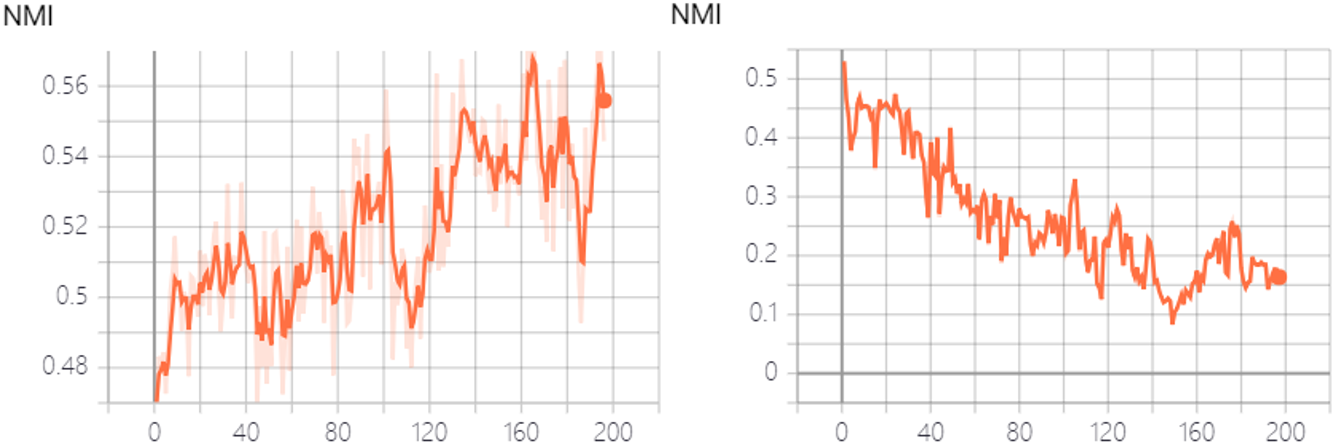}}
\end{minipage}
}

\caption{On the left of the plots (a) to (c) are the performance results of DAFC, which are Acc, ARI, and NMI on the \textbf{F-MNIST}, respectively. On the right of the plots (a) to (c) are the performance results of DAC.}
\end{figure}

We have trained our method and DAC with optimal parameters and obtain the values of Acc, ARI, and NMI, showing each result of cluster assignments for training F-MNIST in Figs. 7(a), 7(b), and 7(c). From the plots, it can be seen that among the two deep clustering methods, the optimal results of Acc, ARI, and NMI are all attained by DAFC with the top-1 values being 0.6180, 0.4862, and 0.5889, respectively. Our method improves more than 7.39\%, 8.90\%, and 7.50\% compared to the DAC on this dataset.

\noindent{\textbf{SVHN}}

\begin{figure}[htbp]
\centering
\subfigure[{Acc values (DAFC and DAC)}]
{
\begin{minipage}[b]{0.2\textwidth}
\centerline{\includegraphics[height=3.4cm, width=9cm]{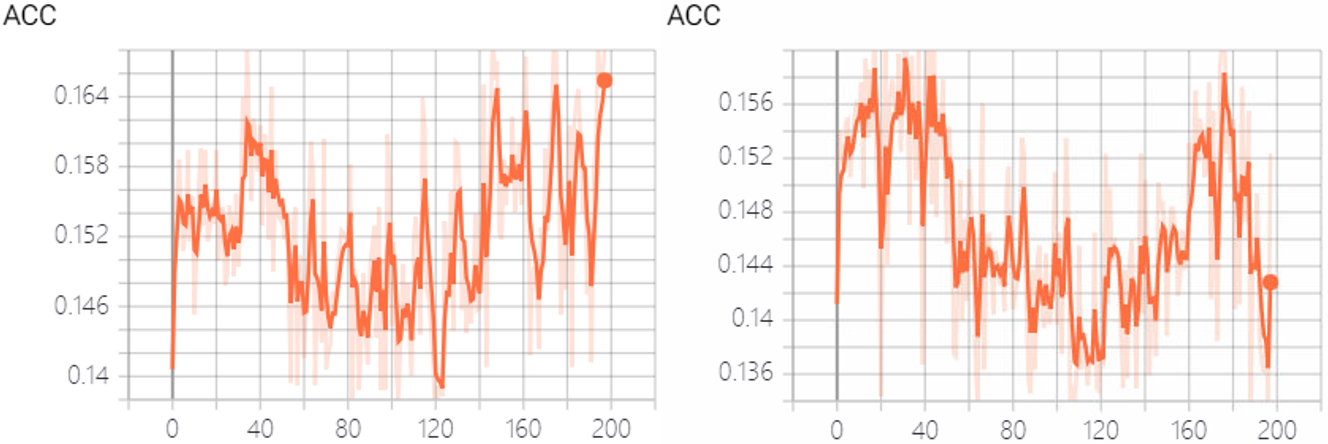}}
\end{minipage}
}

\subfigure[ARI values (DAFC and DAC)]{
\begin{minipage}[b]{0.2\textwidth}
\centerline{\includegraphics[height=3.4cm, width=9cm]{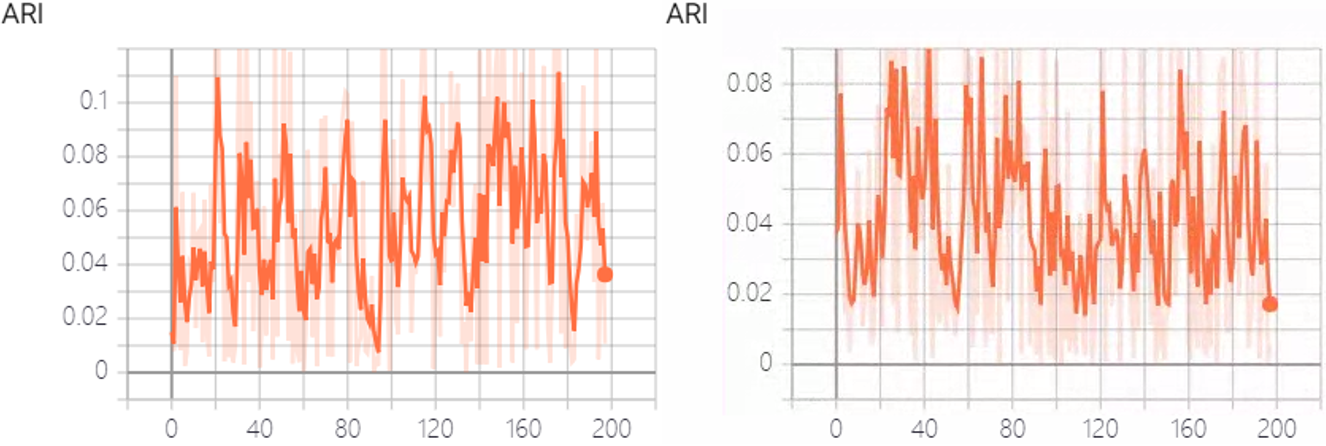}}
\end{minipage}
}

\subfigure[NMI values (DAFC and DAC)]{
\begin{minipage}[b]{0.2\textwidth}
\centerline{\includegraphics[height=3.4cm, width=9cm]{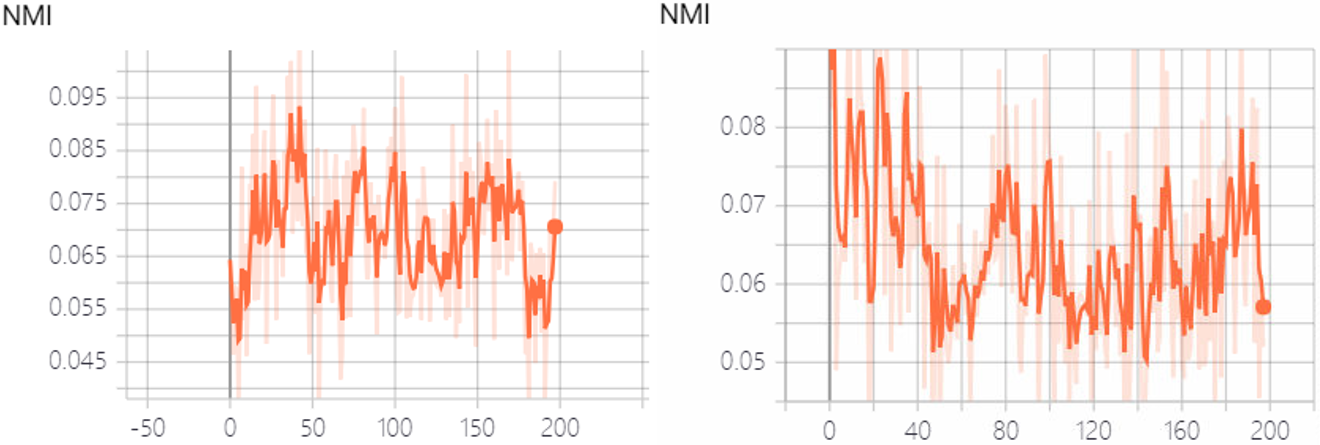}}
\end{minipage}
}

\caption{On the left of the plots (a) to (c) are the performance results of DAFC, which are Acc, ARI, and NMI on the \textbf{SVHN}, respectively. On the right of the plots (a) to (c) are the performance results of DAC.}
\end{figure}

Under the same environment and settings, we have also trained DAFC and DAC with optimal parameters and obtain the best values of Acc, ARI, and NMI, showing each result of cluster assignments for training SVHN in Figs. 8(a), 8(b), and 8(c). From the above plots, it can be seen that among the two deep clustering methods, the optimal results of Acc, ARI, and NMI are attained by DAFC with the best values being 0.1701, 0.1431, and 0.1149, respectively. Our method improves more than 0.51\%, 0.53\%, and 0.15\% compared to the DAC on SVHN for three metrics. As with the MNIST and F-MNIST, the proposed method yields better overall performance than the comparison method.

\subsubsection{Comparison with DCCM}

Further analysis, we show for each training and cluster assignment results of the DCCM\footnote{https://github.com/Cory-M/DCCM} method and our method the following: Acc, ARI, and NMI on the CIFAR-10, CIFAR-100, and STL-10 datasets. By observing and comparing these results, we can find which method is better.

\noindent{\textbf{CIFAR-10}}

\begin{figure}[htbp]
\centering
\subfigure[Acc for DAFC and DCCM]
{
\begin{minipage}[b]{0.2\textwidth}
\centerline{\includegraphics[height=3.4cm, width=9cm]{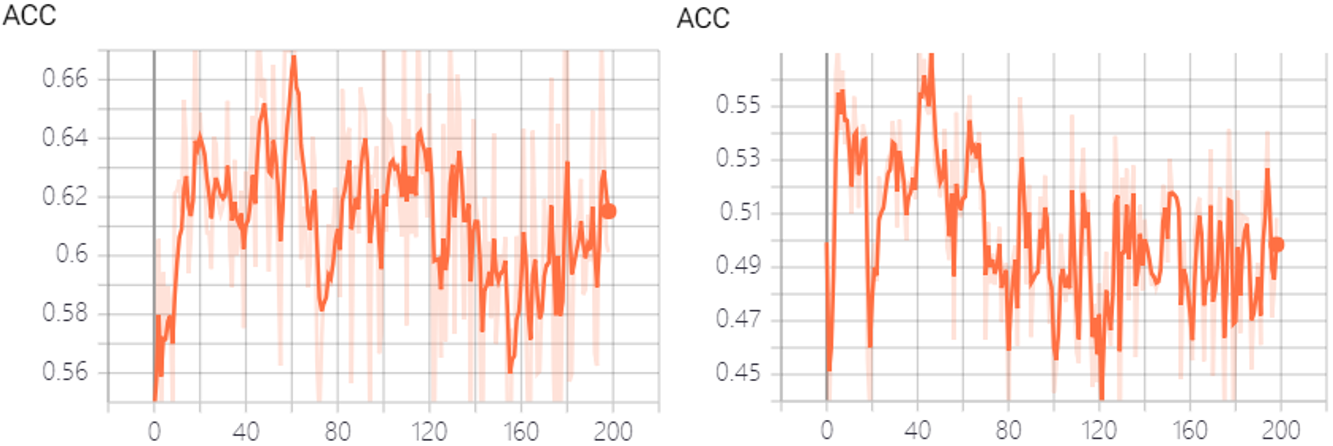}}
\end{minipage}
}

\subfigure[ARI for DAFC and DCCM]{
\begin{minipage}[b]{0.2\textwidth}
\centerline{\includegraphics[height=3.4cm, width=9cm]{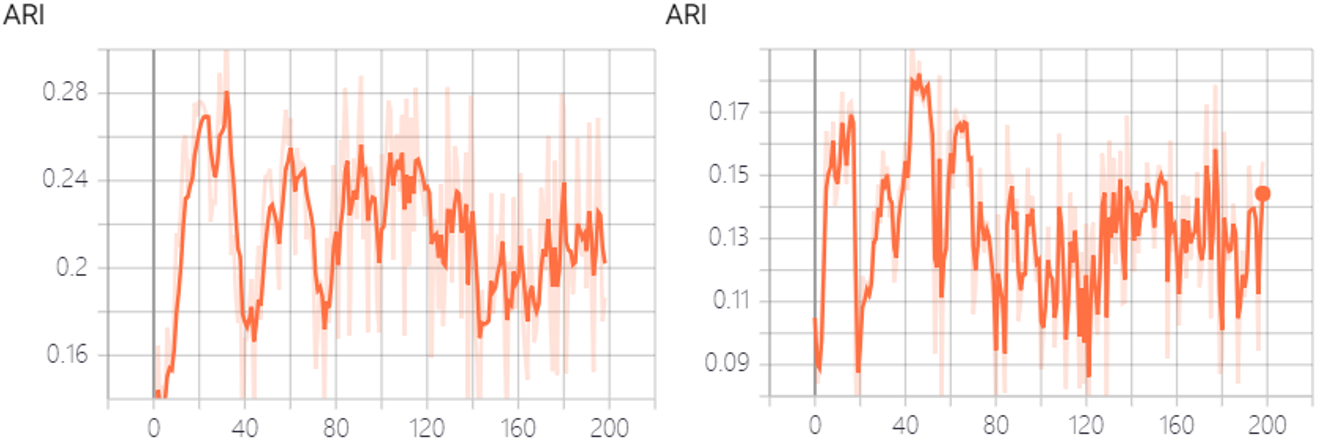}}
\end{minipage}
}

\subfigure[NMI for DAFC and DCCM]{
\begin{minipage}[b]{0.2\textwidth}
\centerline{\includegraphics[height=3.4cm, width=9cm]{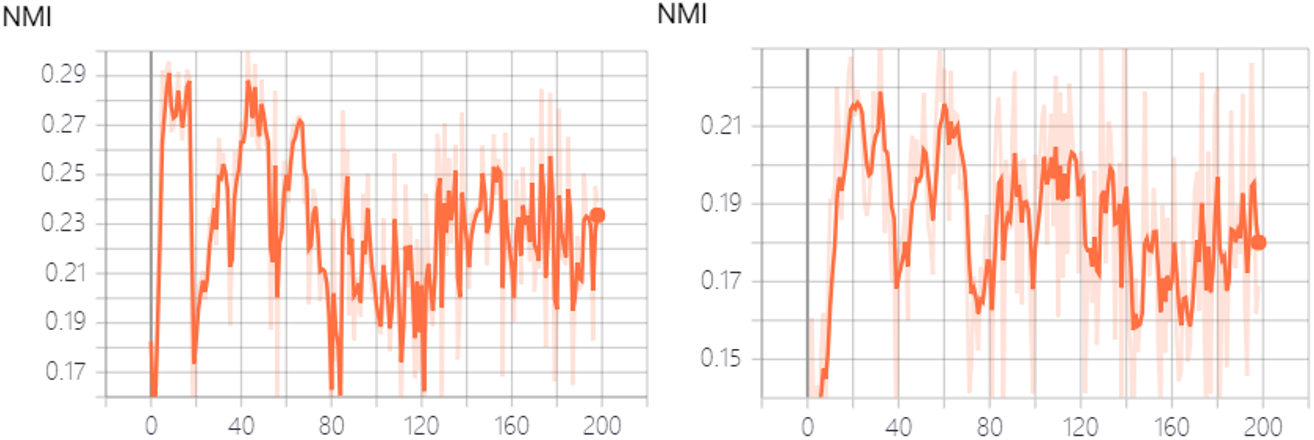}}
\end{minipage}
}

\caption{On the left of the plots (a) to (c) are the performance results of DAFC, which are Acc, ARI, and NMI on the \textbf{CIFAR-10}, respectively. On the right of the plots (a) to (c) are the performance results of DCCM.}
\end{figure}

Further training our method and comparison method (DCCM) with optimal parameters on this dataset. For DAFC, obtaining the top-1 Acc, ARI, and NMI are 0.6876, 0.2912, and 0.3025, respectively. For DCCM, obtaining the top-1 Acc, ARI, and NMI are 0.6230, 0.2280, and 0.2410, respectively. We show the distribution of cluster assignment results for training this dataset in Figs. 9(a), 9(b), and 9(c). From these plots, we can find that the proposed method improves more than 6.46\%, 6.38\%, and 6.15\% compared to the DCCM on CIFAR-10, as well as shows good clustering performance.

\noindent{\textbf{CIFAR-100}}

\begin{figure}[htbp]
\centering
\subfigure[{Acc for DAFC and DCCM}]
{
\begin{minipage}[b]{0.2\textwidth}
\centerline{\includegraphics[height=3.4cm, width=9cm]{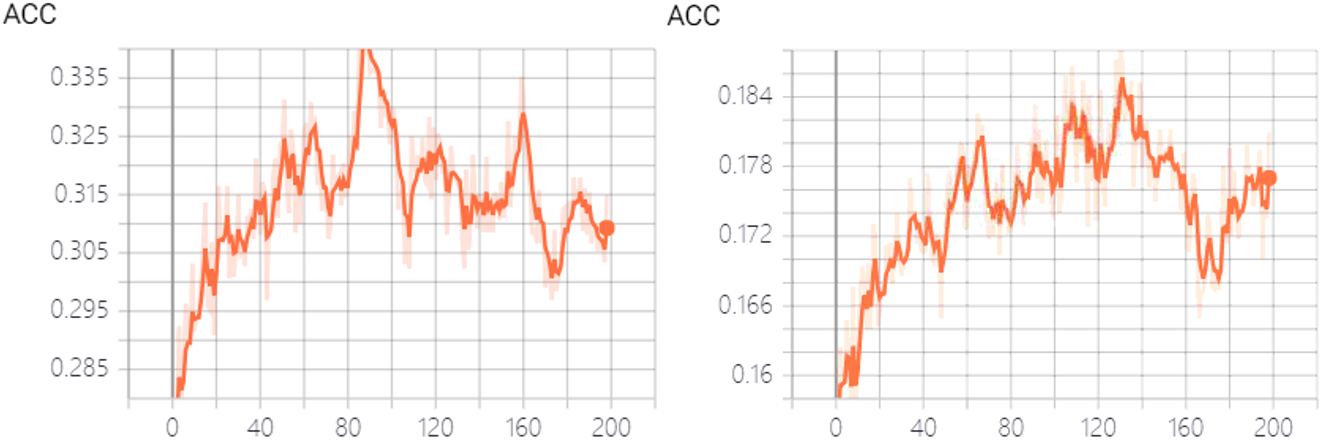}}
\end{minipage}
}

\subfigure[ARI for DAFC and DCCM]{
\begin{minipage}[b]{0.2\textwidth}
\centerline{\includegraphics[height=3.4cm, width=9cm]{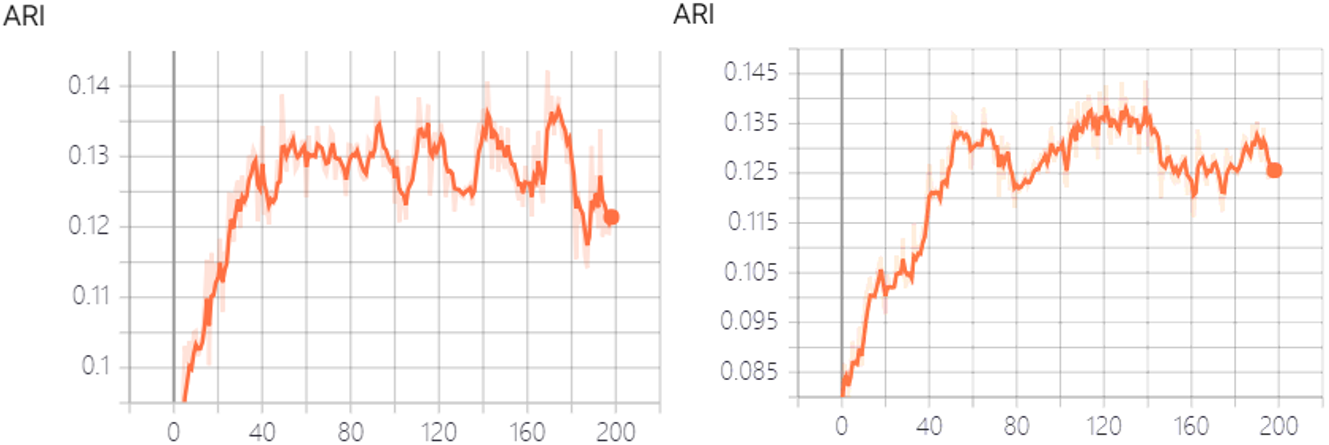}}
\end{minipage}
}

\subfigure[NMI for DAFC and DCCM]{
\begin{minipage}[b]{0.2\textwidth}
\centerline{\includegraphics[height=3.4cm, width=9cm]{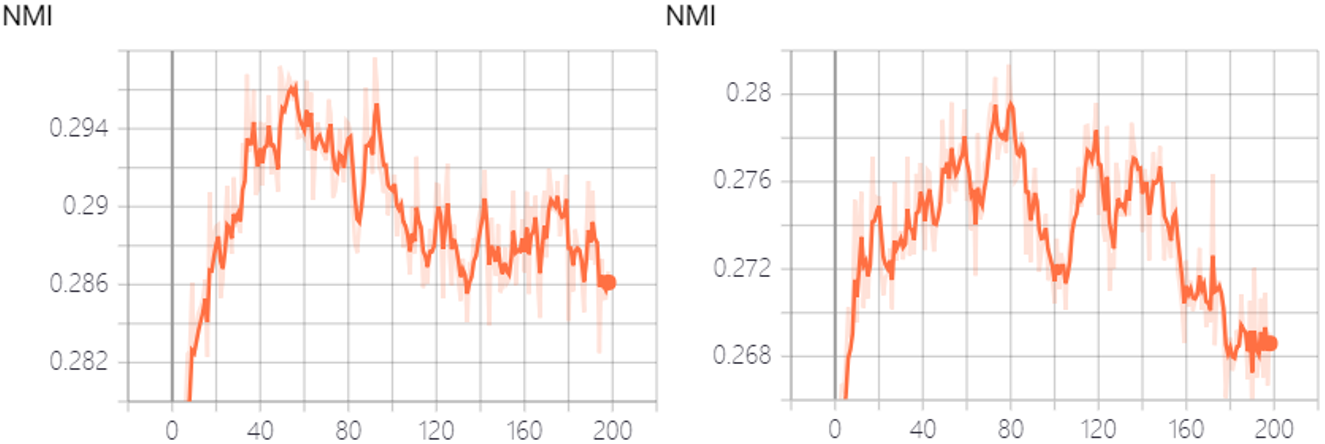}}
\end{minipage}
}

\caption{On the left of the plots (a) to (c) are the performance results of DAFC, which are Acc, ARI, and NMI on the \textbf{CIFAR-100}, respectively. On the right of the plots (a) to (c) are the performance results of DCCM.}
\end{figure}

We have trained our method and DCCM with optimal parameters and obtain the values of Acc, ARI, and NMI, showing each result of cluster assignments for training CIFAR-100 in Figs. 10(a), 10 (b), and 10 (c). From the plots it can be seen that among the two deep clustering methods the optimal results of Acc, ARI, and NMI are all attained by DAFC with the top-1 values being 0.3359, 0.1388, and 0.2976, respectively. For Acc and NMI, our method improves more than 14.71\% and 1.63\% compared to the DCCM on CIFAR-100. But comparing ARI our method is 0.51\% lower than DCCM. Despite the fact that it is not always ranking the first, DAFC, in total, is the best and yields stable performance.

\noindent{\textbf{STL-10}}

\begin{figure}[htbp]
\centering
\subfigure[{Acc for DAFC and DCCM}]
{
\begin{minipage}[b]{0.2\textwidth}
\centerline{\includegraphics[height=3.5cm, width=9cm]{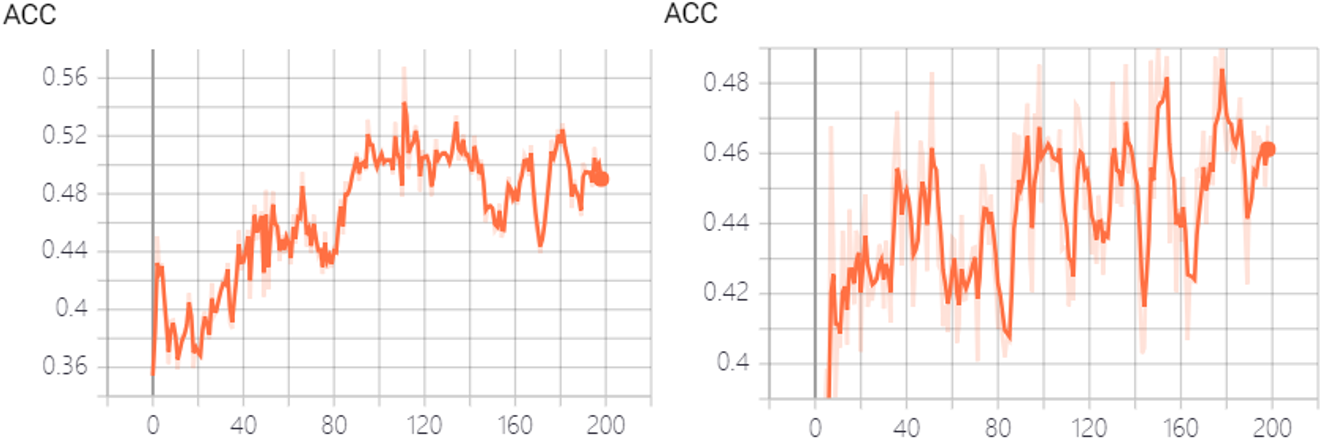}}
\end{minipage}
}

\subfigure[ARI for DAFC and DCCM]{
\begin{minipage}[b]{0.2\textwidth}
\centerline{\includegraphics[height=3.5cm, width=9cm]{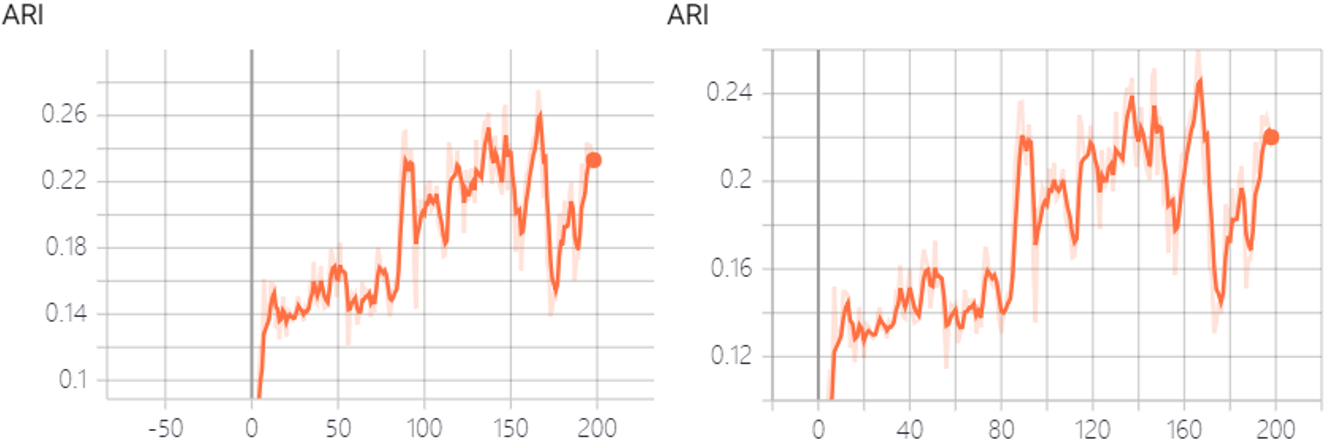}}
\end{minipage}
}

\subfigure[NMI for DAFC and DCCM]{
\begin{minipage}[b]{0.2\textwidth}
\centerline{\includegraphics[height=3.5cm, width=9cm]{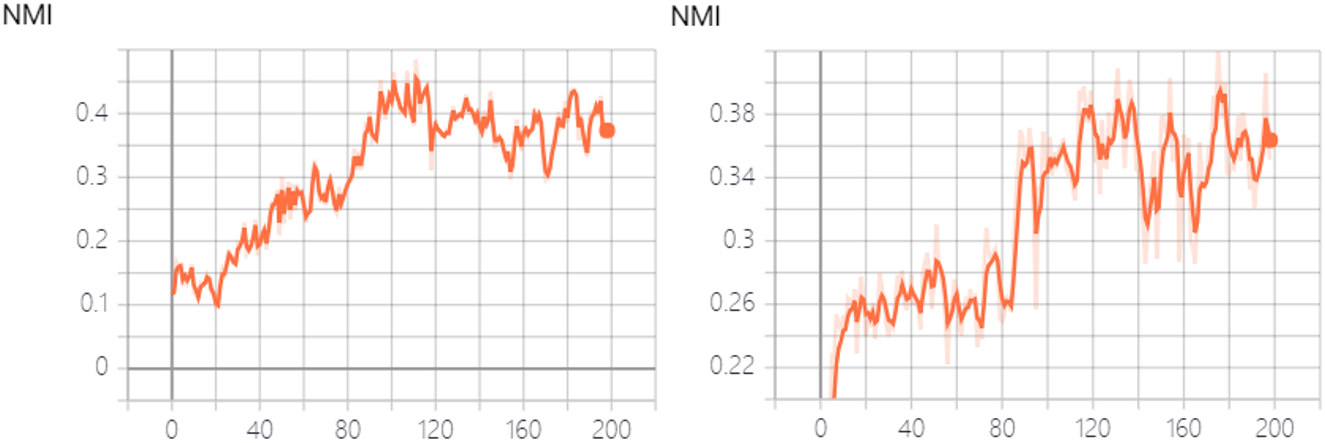}}
\end{minipage}
}

\caption{On the left of the plots (a) to (c) are the performance results of DAFC, which are Acc, ARI, and NMI on the \textbf{STL-10}, respectively. On the right of the plots (a) to (c) are the performance results of DCCM.}
\end{figure}

Under the same environment and settings, we have also trained our method and DCCM with optimal parameters and obtain the best values of Acc, ARI, and NMI, showing each result of cluster assignments for training STL-10 in Figs. 11(a), 11(b), and 11(c). Also, from the plots it can be seen that among the two deep clustering methods the optimal results of Acc, ARI, and NMI are attained by DAFC with the top-1 values being 0.5731, 0.2752, and 0.4260, respectively. Our method improves more than 4.92\%, 1.54\%, and 2.93\% compared to the DCCM on STL-10 for three metrics. In comparisons with DCCM, since more excellent feature representations will be captured during cluster assignments and benefited from the effective representation learning, impressive improvements are achieved and learned by our method.

As for the previous experiments, we outperform previous deep clustering methods on all three metrics. The results imply that DAFC achieves state-of-the-art performance on the deep clustering task. 

\subsubsection{Comparison with state-of-the-art deep clustering}

In order to obtain a more comprehensive assessment of DAFC, we have compared its clustering performance with the selected baselines, which are widely applied in various applications. In Tables I, II, and III, we present the quantitative cluster assignment results of the comparison methods, and they summarized the comparisons of traditional clustering methods and deep clustering with other feature learning approaches on the three metrics. 

\begin{table*}[htbp]
\centering
\caption{Acc of different approaches on six challenging datasets. The highlighted numbers in \textbf{bold} which represent the best results.}
\renewcommand\arraystretch{1.5}
\begin{tabular}{c|cccccccccccc}
\hline
Datasets\textbackslash{}Methods & K-means & AC     & SC     & DeCNN  & DAE    & VAE    & GAN    & JULE   & DEC    & DAC    & DCCM   & DAFC   \\ \hline
MNIST                           & 0.4423  & 0.4753 & 0.4793 & 0.7236 & 0.8129 & 0.8213 & 0.8176 & 0.9336 & 0.8330 & 0.9461 & 0.9578 & \textbf{0.9703} \\
F-MNIST                         & 0.2255  & 0.3491 & 0.3591 & 0.3933 & 0.4942 & 0.5053 & 0.4957 & 0.5343 & 0.5163 & 0.5441 & 0.5739 & \textbf{0.6180}\\
SVHN                            & -       & -      & -      & 0.0926 & 0.1090 & 0.1095 & 0.1120 & 0.1521 & 0.1372 & 0.1650 & 0.1491 & \textbf{0.1701}\\
CIFAR-10                        & 0.2113  & 0.2275 & 0.2467 & 0.2820 & 0.2673 & 0.2830 & 0.3074 & 0.2715 & 0.3011 & 0.5218 & 0.6230 & \textbf{0.6876} \\
CIFAR-100                       & 0.1197  & 0.1378 & 0.1360 & 0.1327 & 0.1405 & 0.1513 & 0.1439 & 0.1367 & 0.1860 & 0.2375 & 0.1888 & \textbf{0.3359} \\
STL-10                          & 0.1752  & 0.3322 & 0.1588 & 0.2855 & 0.2536 & 0.2792 & 0.2765 & 0.2679 & 0.3490 & 0.4899 & 0.5039 & \textbf{0.5631} \\ \hline
\end{tabular}
\end{table*}

\begin{table*}[htbp]
\centering
\caption{ARI of different approaches on six challenging datasets. The highlighted numbers in \textbf{bold} which represent the best results.}
\renewcommand\arraystretch{1.5}
\begin{tabular}{c|cccccccccccc}
\hline
Datasets\textbackslash{}Methods & K-means & AC     & SC     & DeCNN  & DAE    & VAE    & GAN    & JULE   & DEC    & DAC    & DCCM   & DAFC   \\ \hline
MNIST                           & 0.2733  & 0.3904 & 0.4933 & 0.6391 & 0.6233 & 0.6141 & 0.7299 & 0.9194 & 0.7414 & 0.9133 & 0.9153 & \textbf{0.9380} \\
F-MNIST                         & 0.1536  & 0.1526 & 0.2199 & 0.4361 & 0.4134 & 0.3958 & 0.5335 & 0.6059 & 0.5397 & 0.3972 & 0.4235 & \textbf{0.4862} \\
SVHN                            & -       & -      & -      & 0.0730 & 0.0815 & 0.0866 & 0.0897 & 0.1135 & 0.1056 & 0.1378 & 0.1139 & \textbf{0.1431} \\
CIFAR-10                        & 0.0937  & 0.0658 & 0.0899 & 0.1836 & 0.1714 & 0.1522 & 0.1852 & 0.1633 & 0.1607 & 0.2159 & 0.2280 & \textbf{0.2912} \\
CIFAR-100                       & 0.0347  & 0.0322 & 0.0314 & 0.0522 & 0.0560 & 0.0507 & 0.0553 & 0.0432 & 0.0595 & 0.0876 & \textbf{0.1473} & 0.1422 \\
STL-10                          & 0.0752  & 0.1211 & 0.0579 & 0.1621 & 0.1677 & 0.1433 & 0.1479 & 0.1899 & 0.1981 & 0.2532 & 0.2598 & \textbf{0.2752} \\ \hline
\end{tabular}
\end{table*}

\begin{table*}[htbp]
\centering
\caption{NMI of different approaches on six challenging datasets. The highlighted numbers in \textbf{bold} which represent the best results.}
\renewcommand\arraystretch{1.5}
\begin{tabular}{c|cccccccccccc}
\hline
Datasets\textbackslash{}Methods & K-means & AC     & SC     & DeCNN  & DAE    & VAE    & GAN    & JULE   & DEC    & DAC    & DCCM   & DAFC   \\ \hline
MNIST                           & 0.3672  & 0.5031 & 0.6211 & 0.7355 & 0.7463 & 0.7524 & 0.7655 & 0.9022 & 0.7933 & 0.8944 & 0.9012 & \textbf{0.9370} \\
F-MNIST                         & 0.1711  & 0.2311 & 0.2756 & 0.5239 & 0.5324 & 0.4973 & 0.5876 & 0.6251 & 0.5877 & 0.5139 & 0.5433 & \textbf{0.5889} \\
SVHN                            & -       & -      & -      & 0.0912 & 0.0976 & 0.0983 & 0.0959 & 0.1105 & 0.1110 & 0.1134 & 0.1117 & \textbf{0.1149} \\
CIFAR-10                        & 0.0871  & 0.1033 & 0.1007 & 0.2295 & 0.2407 & 0.2551 & 0.2599 & 0.2032 & 0.2667 & 0.2359 & 0.2410 & \textbf{0.3025} \\
CIFAR-100                       & 0.0839  & 0.0976 & 0.0903 & 0.0956 & 0.1139 & 0.1183 & 0.1233 & 0.1126 & 0.1390 & 0.1852 & 0.2813 & \textbf{0.2976} \\
STL-10                          & 0.1143  & 0.2355 & 0.0988 & 0.2339 & 0.2311 & 0.2662 & 0.2177 & 0.1917 & 0.2631 & 0.3656 & 0.4260 & \textbf{0.4553} \\ \hline
\end{tabular}
\end{table*}

From Tables I, II, and III, it can be seen that the performance of the deep feature presentation-based clustering methods (e.g., DeCNN, DAE, and VAE) is superior to the traditional clustering methods (e.g., K-means, AC, and SC) currently. It shows that the clustering techniques have only a minor impact on performance, while the (deep) feature representation learning is more crucial than without feature representation.

For each metric, our method achieves the best results in all the six challenging datasets. From comparison with the best results of the baselines, our deep clustering obtains a significant improvement of 3.44\% on Acc, 3.97\% on ARI, and 3.97\% on NMI averagely when compared to DAC. A significant improvement of 8.69\% on Acc, 2.47\% on ARI, and 3.57\% on NMI averagely when compared to DCCM. In particular, dramatical superiorities are obtained by DAFC on the CIFAR-100, SVHN, and STL-10, which signifies that DAFC can be utilized on large-scale and complex datasets, not merely limited to some simple datasets.

\subsection{Discussions}
\subsubsection{Impact of Global Variables}

To investigate the stability of the proposed method and jointly optimize for the deep representation learning and clustering, we compare the performance of DAFC under various global variables $\mu$ and $w$. Also, in order to reduce the hyperparameter search, we uniformly set the local balance coefficients $\lambda$ to 0.50. From Figs. 12 (a), (b), and (c), it can be seen that the global variables have a significant influence on the performance of DAFC, and the performance distributions are shown in these plots. Since our deep ConvNets with a low learning rate is initialized randomly, more outlier labeled pairwise patterns may be employed for training if $\mu$ and $w$ are too small.

\begin{figure}[htbp]
\centering
\subfigure[The distributions of Acc]{
\begin{minipage}[b]{0.45\textwidth}
\centerline{\includegraphics[height=4.5cm, width=9cm]{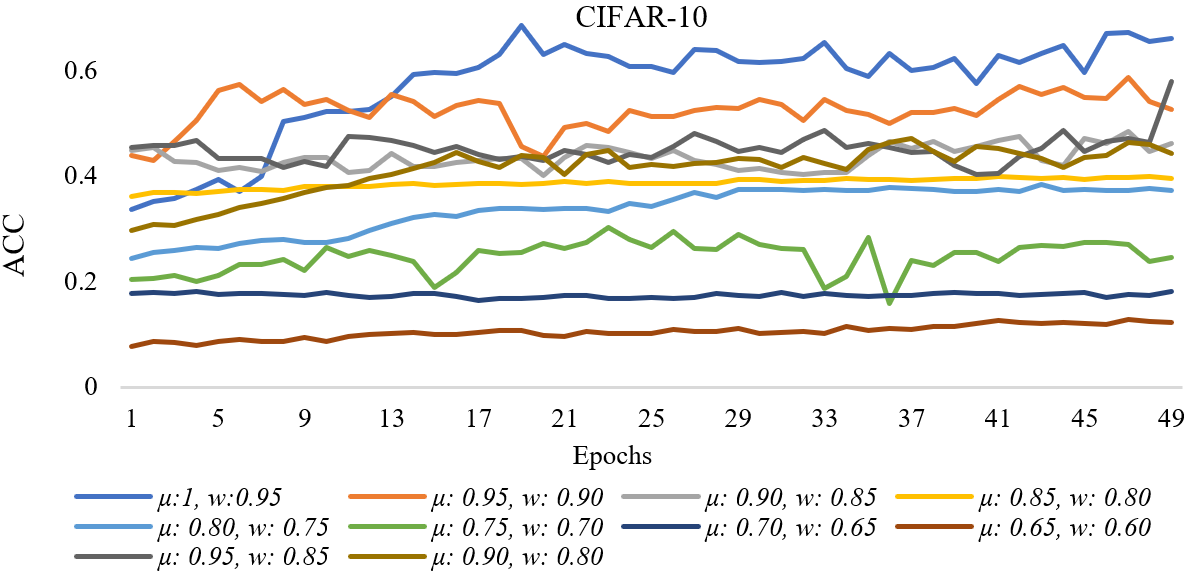}}
\end{minipage}
}

\subfigure[The distributions of ARI]{
\begin{minipage}[b]{0.45\textwidth}
\centerline{\includegraphics[height=4.5cm, width=9cm]{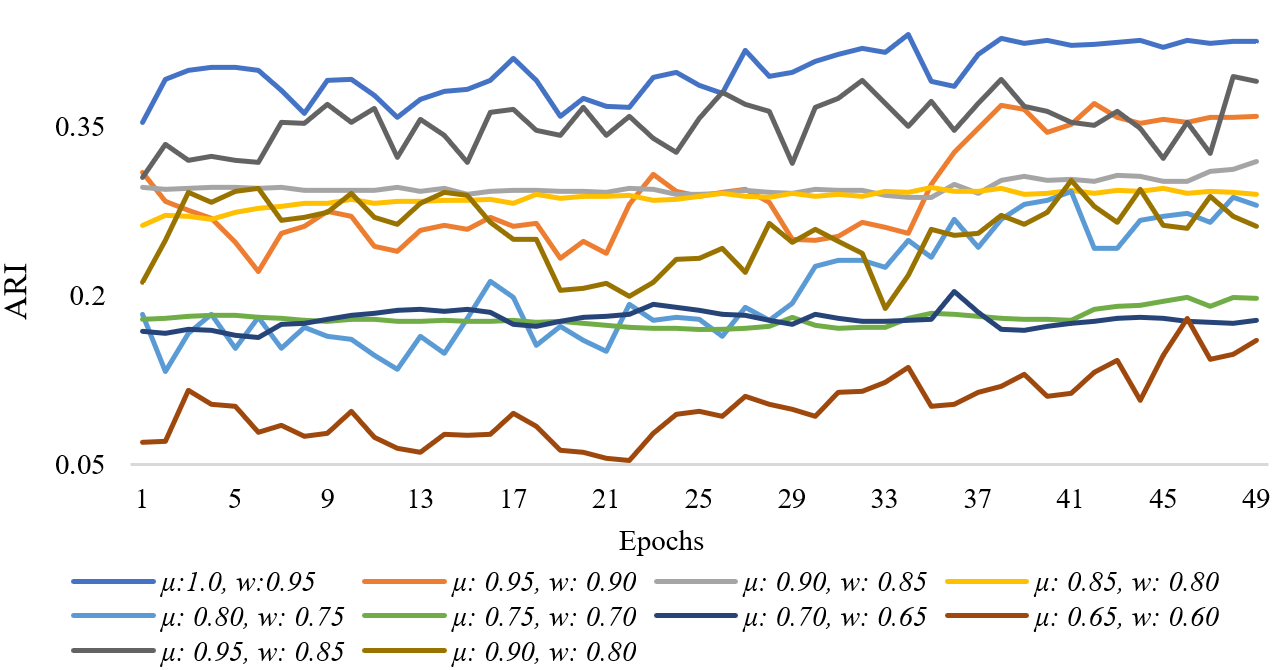}}
\end{minipage}
}
\subfigure[The distributions of NMI]{
\begin{minipage}[b]{0.45\textwidth}
\centerline{\includegraphics[height=4.5cm, width=9cm]{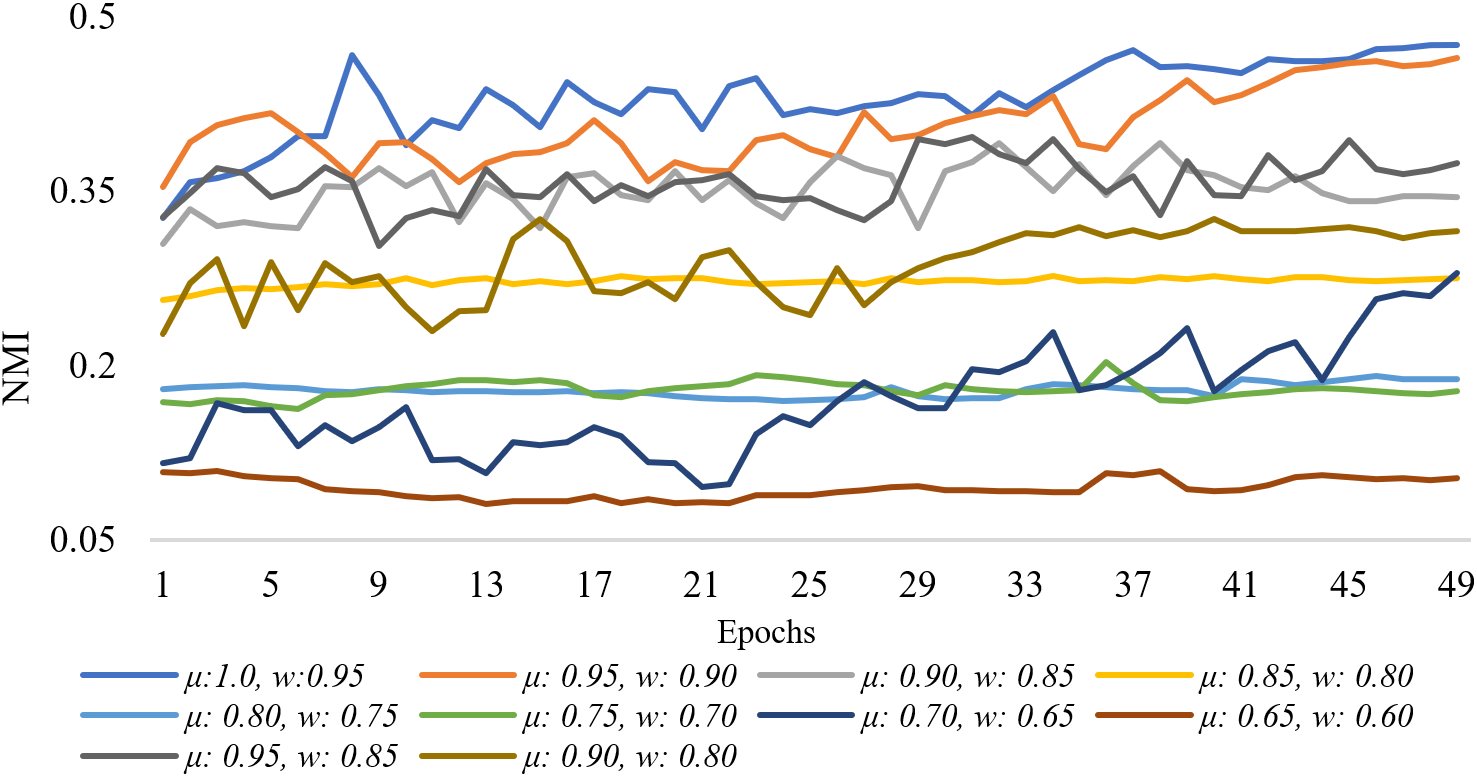}}
\end{minipage}
}

\caption{DAFC with the different global variables of $\mu$ and $w$.}
\end{figure}

From the initialization of hyperparameters, more appropriate values can help DAFC to select highly-confident generated label features. Through the joint fuzzy clustering in the deep neural network with the high global threshold values, DAFC can begin with more refined cluster assignments and yields the better clustering finally.

\subsubsection{Contribution of Deep ConvNets}

As mentioned in the introduction, what types of neural networks are proper for feature representation learning of clustering? In order to explore the contribution of deep ConvNets in the feature representation learning of clustering, Figs. 13 to 18 exhibit the distributions of the proposed method on six highly competitive datasets. An excellent deep clustering model should have both good clustering performance and ConvNets training model, otherwise the output may be unsatisfactory due to the freedom of the joint deep ConvNet model. Under this scenario, we further train and analyze the test error rates of the model and the accuracy of clustering simultaneously and present a more comprehensive assessment of DAFC. In order to keep the figures readable, we also show the test error rates of DAFC from Figs. 13 to 18 on six challenging datasets, which are summarized in Table IV.

\begin{figure}[!htbp]
\centering
\subfigure{
\begin{minipage}{0.53\linewidth}
\centering
\centerline{\includegraphics[height=4.5cm,width=4.9cm]{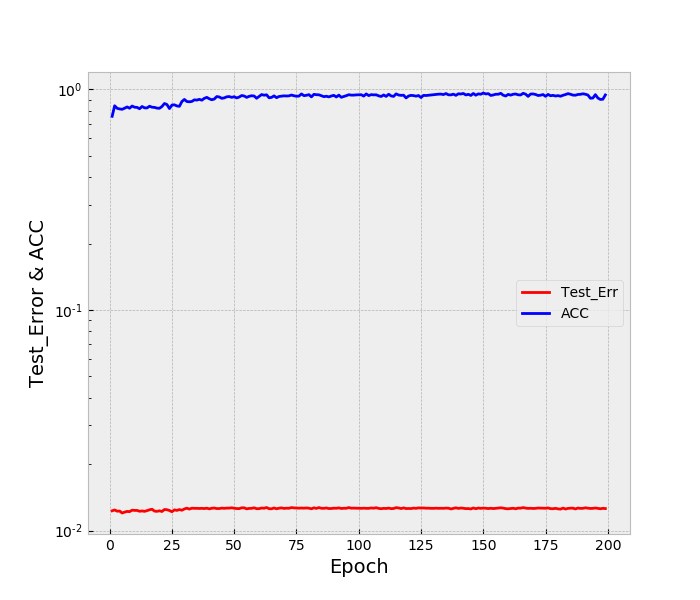}}
\end{minipage}%
}%
\subfigure{
\begin{minipage}{0.45\linewidth}
\centering
\centerline{\includegraphics[height=4.5cm,width=4.9cm]{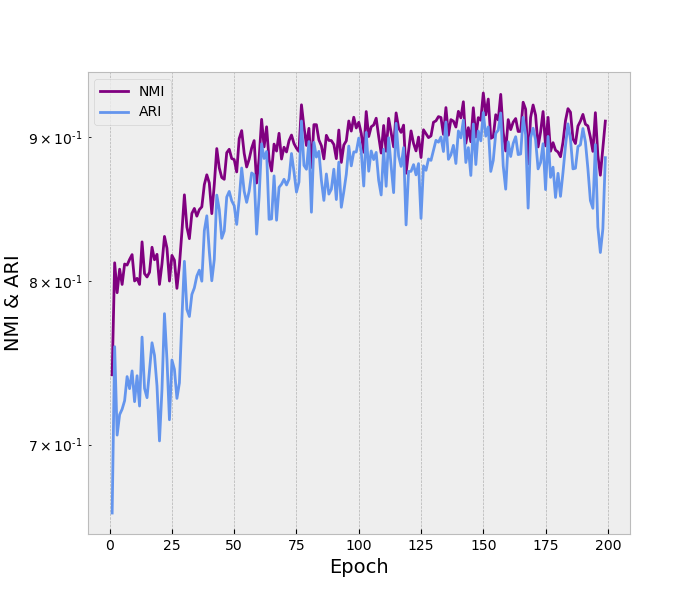}}
\end{minipage}%
}
\caption{Clustering and deep ConvNets performance distributions of DAFC on \textbf{MNIST}.}
\end{figure}

Comparing the results in Fig. 13 to 18, we can find that no matter which types of datasets, DAFC achieves the best accuracy of clustering and test-error rates simultaneously. The ARI and NMI are increasing, meaning that there are less and less cluster reassignments are stabilizing over time. Moreover, the saturation values of ARI and NMI keep changing horizontally, meaning that a significant fraction of data samples are regularly reassigned between epochs, which are the best clustering state along with the iterative optimization. These prove that both deep FE and Rec models play an important role in improving the fuzzy clustering quality of representation learning. 

\begin{figure}[!htbp]
\centering
\subfigure{
\begin{minipage}{0.50\linewidth}
\centering
\centerline{\includegraphics[height=4.5cm,width=4.9cm]{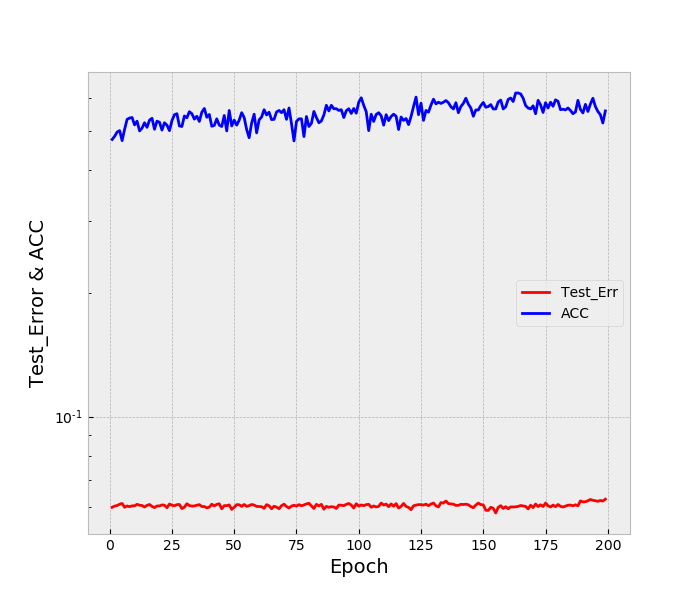}}
\end{minipage}%
}%
\subfigure{
\begin{minipage}{0.460\linewidth}
\centering
\centerline{\includegraphics[height=4.5cm,width=4.9cm]{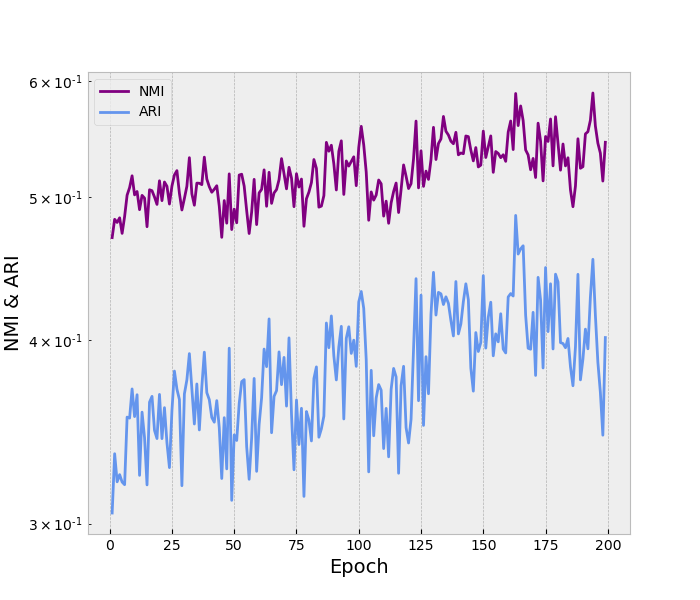}}
\end{minipage}%
}
\caption{Clustering and deep ConvNets performance distributions of DAFC on \textbf{F-MNIST}.}
\end{figure}

\begin{figure}[!htbp]
\centering
\subfigure{
\begin{minipage}{0.52\linewidth}
\centering
\centerline{\includegraphics[height=4.5cm,width=5cm]{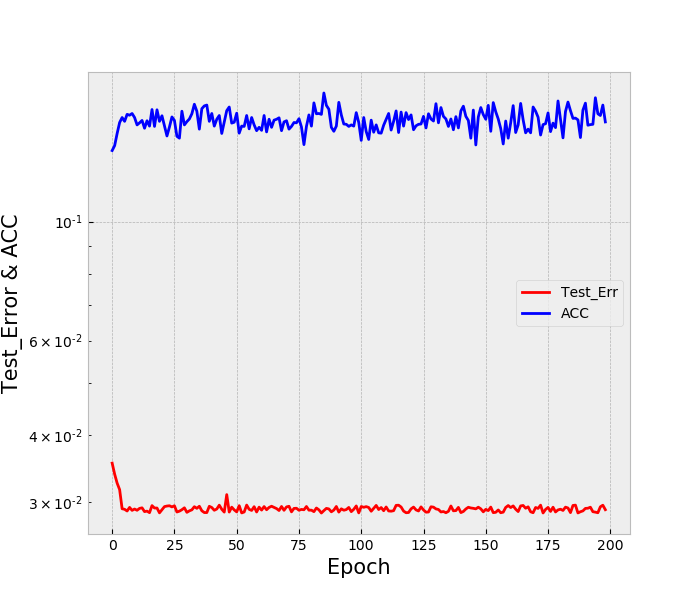}}
\end{minipage}%
}%
\subfigure{
\begin{minipage}{0.460\linewidth}
\centering
\centerline{\includegraphics[height=4.5cm,width=5cm]{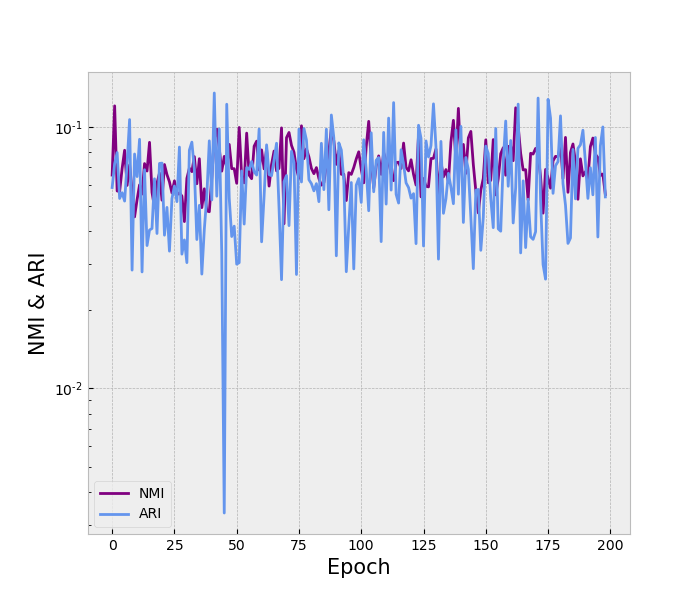}}
\end{minipage}%
}
\caption{Clustering and deep ConvNets performance distributions of DAFC on \textbf{SVHN}.}
\end{figure}

\begin{figure}[!htbp]
\centering
\subfigure{
\begin{minipage}{0.50\linewidth}
\centering
\centerline{\includegraphics[height=4.5cm,width=4.9cm]{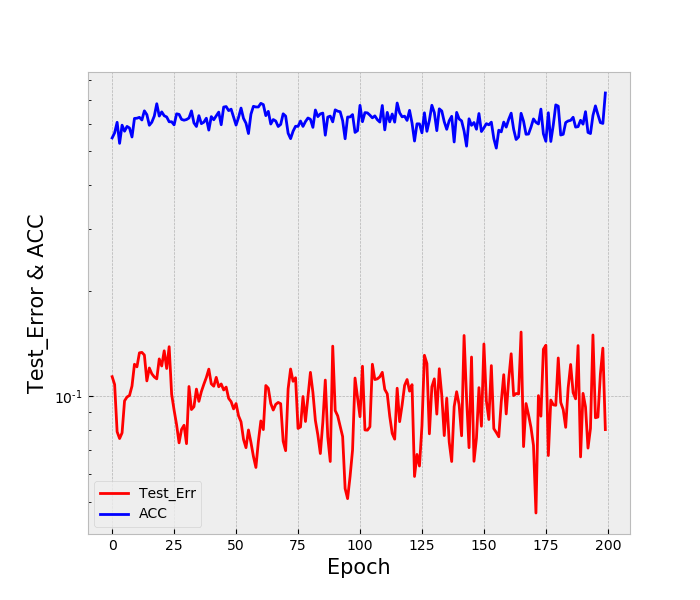}}
\end{minipage}%
}%
\subfigure{
\begin{minipage}{0.460\linewidth}
\centering
\centerline{\includegraphics[height=4.5cm,width=4.9cm]{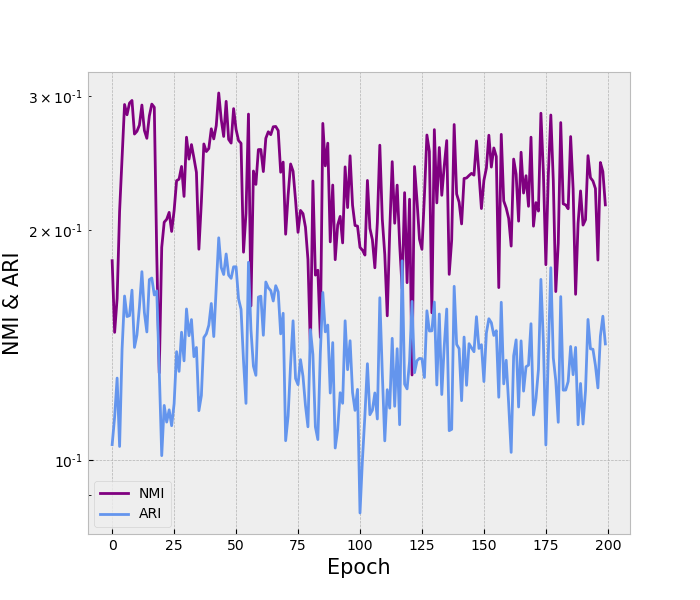}}
\end{minipage}%
}
\caption{Clustering and deep ConvNets performance distributions of DAFC on \textbf{CIFAR-10}.}
\end{figure}

\begin{figure}[!htbp]
\centering
\subfigure{
\begin{minipage}{0.50\linewidth}
\centering
\centerline{\includegraphics[height=4.5cm,width=4.9cm]{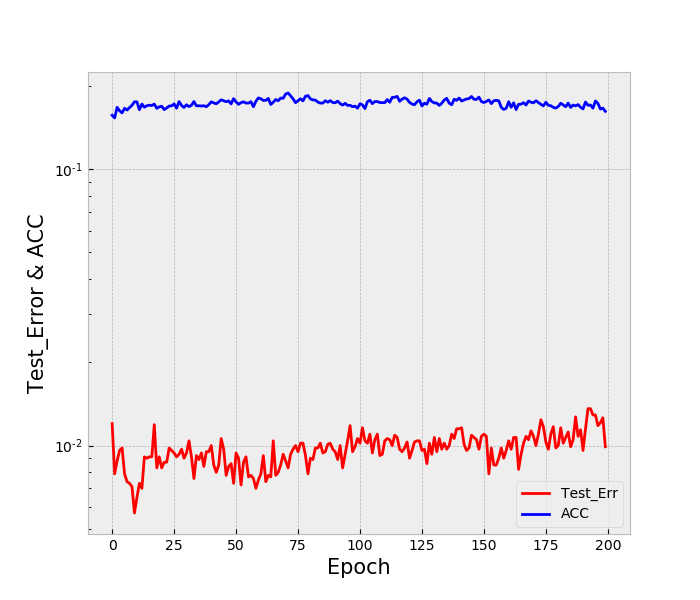}}
\end{minipage}%
}%
\subfigure{
\begin{minipage}{0.460\linewidth}
\centering
\centerline{\includegraphics[height=4.5cm,width=4.9cm]{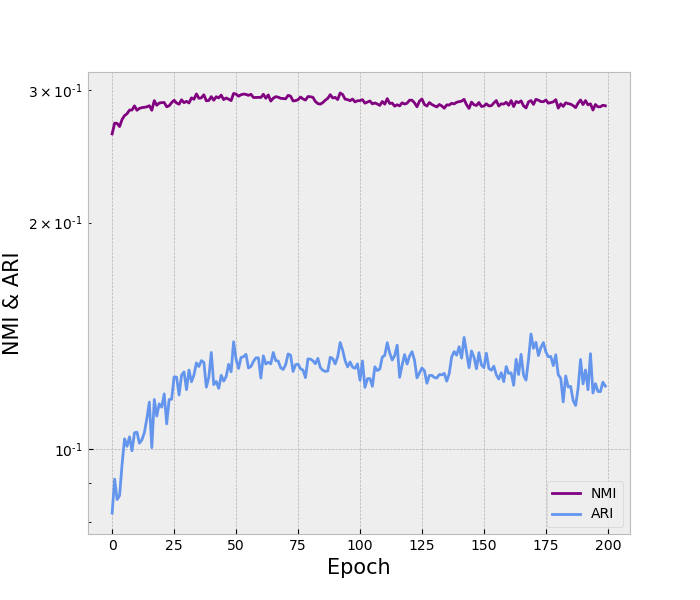}}
\end{minipage}%
}
\caption{Clustering and deep ConvNets performance distributions of DAFC on \textbf{CIFAR-100}.}
\end{figure}

\begin{figure}[!htbp]
\centering
\subfigure{
\begin{minipage}{0.50\linewidth}
\centering
\centerline{\includegraphics[height=4.5cm,width=5.2cm]{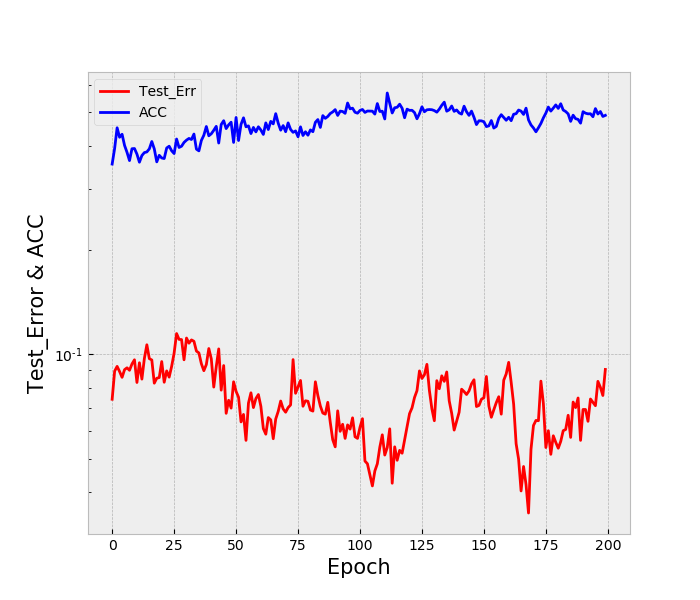}}
\end{minipage}%
}%
\subfigure{
\begin{minipage}{0.460\linewidth}
\centering
\centerline{\includegraphics[height=4.5cm,width=5cm]{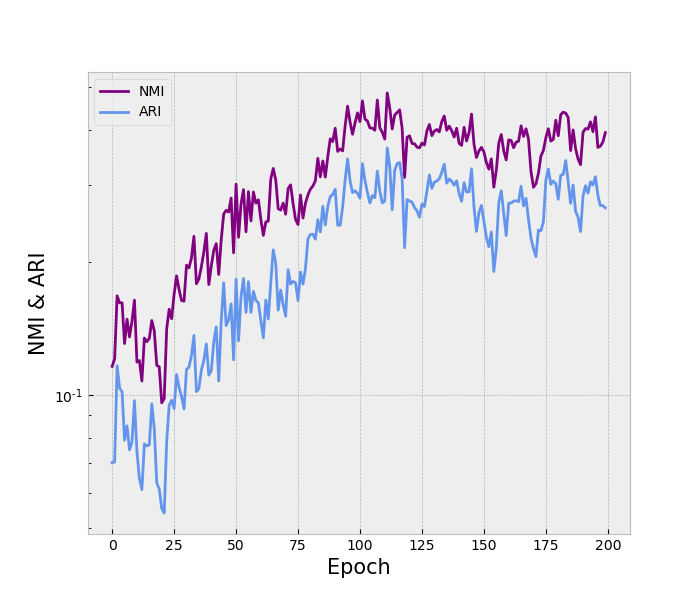}}
\end{minipage}%
}
\caption{Clustering and deep ConvNets performance distributions of DAFC on \textbf{STL-10}.}
\end{figure}

\begin{table}[htbp]
\centering
\caption{Test-Error rates (\%) of different deep clustering frameworks on six challenging datasets.}
\renewcommand\arraystretch{1.6}
\begin{tabular}{c|ccc}
\hline
\multirow{2}{*}{Datasets} & \multicolumn{3}{c}{Test Error Rates} \\ \cline{2-4} 
                          & Our ConvNets             & ResNets          & VGG-16  \\ \hline 
MNIST                     & 1.20             & 4.91      & \textbf{0.67}        \\
F-MNIST                   & \textbf{5.81}             & 5.90      & 6.05        \\
SVHN                      & 2.90             & \textbf{2.01}      & 12.70        \\
CIFAR-10                  & \textbf{5.87}             & 8.31      & 9.19        \\
CIFAR-100                 & \textbf{0.69}             & 27.82     & 16.18       \\
STL-10                    & \textbf{4.03}             & -               & 14.37  \\ \hline
\end{tabular}
\end{table}

In the deep neural network setting, deeper architectures like ResNet have a much higher accuracy on highly competitive datasets than VGG-16 [42] and AlexNet. We should expect the same improvement if ResNet architectures are used with an unsupervised feature representation approach. Table IV describes the proposed method and a ResNet framework trained with deep clustering on six challenging datasets. We observe that regardless of approach, a deeper architecture leads to a significant improvement in clustering performance on the cluster assignments.

Taking the above discussions into consideration, the deep ConvNet model with high accuracy provides the best guarantee for the performance of deep clustering. Furthermore, the superior performance reason is that DAFC successfully integrates the fuzzy clustering into deep evolutionary unsupervised representation learning, and the dual ConvNets model guides the update of the deep clustering model.

\section{Conclusions}

We have presented deep adaptive fuzzy clustering for the evolutionary unsupervised representation learning model, a novel deep feature representation learning is proposed for cluster assignments in this study. We have explored the possibility of employing fuzzy clustering in a deep neural network. The proposed method implemented the deep fuzzy clustering strategy that learns a convolutional neural network classifier from given six challenging datasets. In the experiments, we only consider for comparison with the state-of-the-art and highly competitive datasets, and present thorough experiments that demonstrate the efficacy and value of the proposed technique evaluated on a wide range of use cases and deeper clustering structures. Furthermore, we compare various clustering performance metrics on these highly competitive datasets used by the existing algorithms. It can effectively achieve significant improvement over the state-of-the-art deep clustering methods.

In this way, an excellent theoretical basis has been laid in the data of industrial processes and in which industrial datasets have many correlations and high-dimensional features, so such data characteristics are very complicated. Next, we will use the industrial datasets to test the performance of our deep clustering model in practical applications and solve the training problem too complicated on the deeper network in the process of dividing industrial data. To this end, we will automatically optimize the hyperparameters of the loss function for the deep fuzzy clustering method and exploring more advanced deep convolutional networks in future works.


\section*{Acknowledgments}

The authors would like to thank McMaster University School of Engineering Practice and Technology for providing the use of workstation and working environment. In addition, we would also like to thank Mahir Jalanko and Chiamaka Okeke of Neural Networks Team for helpful comments during the course of this work.

\ifCLASSOPTIONcaptionsoff
  \newpage
\fi





\begin{thebibliography}{1}

\bibitem{11-1}
M.~Caron, P.~Bojanowski, A.~Joulin, and M.~Douze, ``Deep clustering for unsupervised learning of visual features," In Proc. European Conference on Computer Vision (ECCV), pp. 132-149, 2018. 


\bibitem{10-2}
J.~Wang, J.~Wang, J.~Song, X.~Xu, H.~T.~Shen, and S.~Li, ``Optimized cartesian k-means," \emph{IEEE Transactions on Knowledge and Data Engineering}, vol. 27, no. 1, pp. 180-192, 2014.

\bibitem{20-3}
 D. Tan, W. Zhong, C. Jiang, X. Peng, and W. He, ``High-order fuzzy clustering algorithms based on multikernel mean shift," \emph{Neurocomputing}, vol. 385, pp. 63-79, 2020.

\bibitem{21-4}
T. Xia, D. Tao, T. Mei, and Y. Zhang, ``Multiview spectral embedding,'' \emph{IEEE Transactions on Systems, Man, and Cybernetics - Part B: Cybernetics}, vol. 40, no. 6, pp. 1438-1446, 2010.

\bibitem{22-5}
P. Goyal and E. Ferrara, ``Graph embedding techniques, applications, and performance: a survey," \emph{Knowledge-Based Systems}, vol. 151, pp. 78-94, 2018.

\bibitem{14-6}
L.~Yang, N.-M.~Cheung, J. Li, and J. Fang, ``Deep clustering by gaussian mixture variational autoencoders with graph embedding," In \emph{Proc. IEEE International Conference on Computer Vision}, pp. 6440-6449, 2019.

\bibitem{23-7}
F. Nie, G. Cai, J. Li, and X. Li, ``Auto-weighted multi-view learning for image clustering and semi-supervised classification," \emph{IEEE Transactions on Image Processing}, vol. 27, no. 3, pp. 1501-1511, 2018.

\bibitem{24-8}
S. Hong, J. Choi, J. Feyereisl, B. Han, and L. S. Davis, ``Joint image clustering and labeling by matrix factorization," \emph{IEEE Transactions on Pattern Analysis and Machine Intelligence}, vol. 38, no. 7, pp. 1411-1424, 2016. 

\bibitem{6-9}
J.~Yang, D.~Parikh, and D.~Batra, ``Joint unsupervised learning of deep representations and image clusters," In \emph{Proc. IEEE Conference on Computer Vision and Pattern Recognition}, pp. 5147-5156, 2016.

\bibitem{7-10}
X.~Ji, J.~F.~Henriques, and A.~Vedaldi, ``Invariant information clustering for unsupervised image classification and segmentation," In \emph{Proc. IEEE International Conference on Computer Vision}, pp. 9865-9874. 2019.

\bibitem{38-11}
Y. Chen, J. Li, H. Xiao, X. Jin, S. Yan, and J. Feng, ``Dual path networks," In \emph{Proc. Advances in Neural Information Processing Systems}, pp. 4467-4475, 2017.

\bibitem{39-12}
Y. Zhang, Y. Tian, Y. Kong, B. Zhong, and Y. Fu, ``Identity mappings in deep residual networks," \emph{IEEE Transactions on Pattern Analysis and Machine Intelligence}, online, 2020.

\bibitem{15-13}
X. Guo, X. Liu, E. Zhu, and J. Yin, ``Deep clustering with convolutional autoencoders," In \emph{Proc. International Conference on Neural Information Processing}, pp. 373-382, 2017.

\bibitem{40-14}
W.-A. Lin, J.-C. Chen, C. D. Castillo, and R. Chellappa, ``Deep density clustering of unconstrained faces," In \emph{Proc. IEEE Conference on Computer Vision and Pattern Recognition}, pp. 8128-8137, 2018.

\bibitem{25-15}
D. Bo, X. Wang, C. Shi, M. Zhu, E. Lu, and P. Cui, ``Structural deep clustering network," In \emph{Proc. The Web Conference 2020}, pp. 1400-1410, 2020.

\bibitem{19-16}
S. N. Tran and A. S. A. Garcez, ``Deep logic networks: inserting and extracting knowledge from deep belief networks," \emph{IEEE Transactions on Neural Networks and Learning Systems}, vol. 29, no. 2, pp. 246-258, 2016.

\bibitem{26-17}
K. Chen, J. Hu, and J. He, ``A framework for automatically extracting overvoltage features based on sparse autoencoder," \emph{IEEE Transactions on Smart Grid}, vol. 9, no. 2, pp. 594-604, 2018.

\bibitem{31-18}
N. Zeng, H. Zhang, B. Song, W. Liu, Y. Li, and A. M. Dobaie, ``Facial expression recognition via learning deep sparse autoencoders," \emph{Neurocomputing}, vol. 273, pp. 643-649, 2018.

\bibitem{13-19}
B.~Yang, X.~Fu, N.~D.~Sidiropoulos, and M.~Yang, ``Towards k-means-friendly spaces: Simultaneous deep learning and clustering," In \emph{Proc. International Conference on Machine Learning}, pp. 3861-3870. 2017.

\bibitem{12-20}
J.~Xie, R.~Girshick, and A.~Farhadi, ``Unsupervised deep embedding for clustering analysis," In \emph{Proc. International Conference on Machine Learning}, pp. 478-487, 2016.

\bibitem{1-21}
X.~Zhan, J.~Xie, Z.~Liu, Y. S. Ong, C. C. Joy, ``Online deep clustering for unsupervised representation learning," In \emph{Proc. IEEE Conference on Computer Vision and Pattern Recognition}, pp. 6688-6697, 2020.

\bibitem{34-22}
J. Chang, G. Meng, L. Wang, S. Xiang, and C. Pan, ``Deep self-evolution clustering," \emph{IEEE Transactions on Pattern Analysis and Machine Intelligence}, vol. 42, no. 4, pp. 809-823, 2020.

\bibitem{42-23}
X. Zhang, Y. Tian, R. Chen, and Y. Jin, ``A decision variable clustering-based evolutionary algorithm for large-scale many-objective optimization," \emph{IEEE Transactions on Evolutionary Computation}, vol. 22, no. 1, pp. 97-112, 2016.

\bibitem{35-24}
Y. Li, J. Zhang, J. Zhang, and K. Huang, ``Discriminative learning of latent features for zero-shot recognition," In \emph{Proc. IEEE Conference on Computer Vision and Pattern Recognition}, pp. 7463-7471, 2018.

\bibitem{36-25}
M. Ye and J. Shen, ``Probabilistic structural latent representation for unsupervised embedding," In \emph{Proc. IEEE Conference on Computer Vision and Pattern Recognition}, pp. 5457-5466, 2020.

\bibitem{29-26}
Y. Fang, Z. Chen, W. Lin, and C. W. Lin, ``Saliency detection in the compressed domain for adaptive image retargeting," \emph{IEEE Transactions on Image Processing}, vol. 21, no. 9, pp. 3888-3901, 2012.

\bibitem{30-27}
K. Zhang, L. Zhang, and M. H. Yang, ``Fast compressive tracking," \emph{IEEE Transactions on Pattern Analysis and Machine Intelligence}, vol. 36, no. 10, pp. 2002-2015, 2014.

\bibitem{33-28}
X. Yang, C. Deng, F. Zheng, J. Yan, and W. Liu, ``Deep spectral clustering using dual autoencoder network," In \emph{Proc. IEEE Conference on Computer Vision and Pattern Recognition}, pp. 4066-4075. 2019.

\bibitem{32-29}
F. Tian, B. Gao, Q. Cui, E. Chen, and T. Liu, ``Leaning deep representations for graph clustering," In \emph{Proc. 28 AAAI conference on Artificial Intelligence}, 2014. 

\bibitem{16-30}
K. G. Dizaji, A. Herandi, C. Deng, W. Cai, and H. Huang, ``Deep clustering via joint convolutional autoencoder embedding and relative entropy minimization," In \emph{Proc. IEEE International Conference on Computer Vision}, pp. 5736-5745, 2017. 

\bibitem{37-31} 
D. Pathak, P. Krahenbuhl, J. Donahue, T. Darrell, and A. A. Efros, ``Context encoders: Feature learning by inpainting," In \emph{Proc. IEEE Conference on Computer Vision and Pattern Recognition}, pp. 2536-2544, 2016.

\bibitem{27-32} 
K. He, X. Zhang, S. Ren, and J. Sun, ``Deep residual learning for image recognition," In \emph{Proc. IEEE Conference on Computer Vision and Pattern Recognition}, pp. 770-778. 2016.

\bibitem{28-33}
G. Huang, Z. Liu, L. Maaten, and K. Q. Weinberger, ``Densely connected convolutional networks," In \emph{Proc. IEEE Conference on Computer Vision and Pattern Recognition}, pp. 4700-4708, 2017.

\bibitem{17-34}
K. C. Gowda and G. Krishna, ``Agglomerative clustering using the concept of mutual nearest neighbourhood," \emph{Pattern Recognition}, vol. 10, no. 2, pp. 105-112, 1978.

\bibitem{18-35}
L. Z. Manor and P. Perona, ``Self-tuning spectral clustering," In \emph{Proc. Advances in Neural Information Processing Systems}, pp. 1601-1608, 2005.

\bibitem{2-36}
M.~Zeiler, D.~Krishnan, G.~Taylor, and R.~Fergus, ``Deconvolutional Networks," In \emph{Proc. 2010 IEEE Computer Society Conference on Computer Vision and Pattern Recognition}, pp. 2528-2535, 2010.

\bibitem{3-37}
P.~Vincent, H.~Larochelle, Y.~Bengio, P.-A.~Manzagol, and L.~Bottou, ``Stacked denoising autoencoders: Learning useful representations in a deep network with a local denoising criterion," \emph{Journal of Machine Learning Research}, vol. 11, no. 12, 2010.

\bibitem{4-38}
D.~P.~Kingma and M.~Welling, ``Auto-encoding variational bayes," \emph{arXiv preprint arXiv:1312.6114}, 2014.

\bibitem{5-39}
A.~Radford, L.~Metz, and S.~Chintala, ``Unsupervised representation learning with deep convolutional generative adversarial networks," \emph{arXiv preprint arXiv:1511.06434}, 2015.


\bibitem{8-40}
J.~Chang, L.~Wang, G.~Meng, S.~Xiang, and C.~Pan, ``Deep adaptive image clustering," In \emph{Proc. IEEE International Conference on Computer Vision}, pp. 5879-5887, 2017.

\bibitem{9-41}
J.~Wu, K.~Long, F.~Wang, C.~Qian, C.~Li, Z.~Lin, and H.~Zha, ``Deep comprehensive correlation mining for image clustering," In \emph{Proc. IEEE International Conference on Computer Vision}, pp. 8150-8159, 2019.

\bibitem{41-42}
X. Wang, K. He, and A. Gupta, ``Transitive invariance for self-supervised visual representation learning," In \emph{Pro. IEEE International Conference on Computer Vision}, pp. 1329-1338, 2017.






\end{thebibliography}
\end{document}